\pdfoutput=1

\documentclass{article}
\usepackage{ijcai24}

\usepackage{times}
\usepackage{soul}
\usepackage{url}
\usepackage{hyperref}
\hypersetup{hidelinks}
\usepackage[utf8]{inputenc}
\usepackage[small]{caption}
\usepackage{graphicx}
\usepackage{amsmath}
\usepackage{amsthm}
\usepackage{booktabs}
\usepackage{algorithm}
\usepackage{algorithmic}
\usepackage[switch]{lineno}

\usepackage{xcolor}
\usepackage[inkscapelatex=false]{svg}
\definecolor{jhy}{rgb}{0.1, 0.2, 0.7}

\usepackage[toc,page]{appendix}
\usepackage{multirow}
\usepackage{amsfonts}
\usepackage{graphicx}
\usepackage{booktabs}
\usepackage[normalem]{ulem}
\useunder{\uline}{\ul}{}
\definecolor{airforceblue}{rgb}{0.36, 0.54, 0.66}
\definecolor{steelblue2}{RGB}{0,90,158}

\title{X-Light: Cross-City Traffic Signal Control Using Transformer on Transformer as Meta Multi-Agent Reinforcement Learner}

\author{
Haoyuan Jiang$^1$\and
Ziyue Li$^{2, \dagger}$\and
Hua Wei$^{3}$\and
Xuantang Xiong$^4$\and
Jingqing Ruan$^4$\and
Jiaming Lu$^5$\and
Hangyu Mao$^6$\And
Rui Zhao$^7$
\affiliations
$^1$Baidu Inc., China\\
$^2$University of Cologne, Germany\\
$^3$Arizona State University, U.S.A\\
$^4$Institute of Automation, Chinese Academy of Sciences, China\\
$^5$Fudan University, China\\
$^6$Peking University, China\\
$^7$Qingyuan Research Institute, China\\
\emails
jianghaoyuan@zju.edu.cn,
zlibn@wiso.uni-koeln.de.
}

\begin{document}

\maketitle

\begin{abstract}
The effectiveness of traffic light control has been significantly improved by current reinforcement learning-based approaches via better cooperation among multiple traffic lights. However, a persisting issue remains: how to obtain a multi-agent traffic signal control algorithm with remarkable transferability across diverse cities? In this paper, we propose a \textbf{T}ransformer \textbf{on} \textbf{T}ransformer (TonT) model for \textit{cross}-city meta multi-agent traffic signal control, named as X-Light: We input the full Markov Decision Process trajectories, and the Lower Transformer aggregates the states, actions, rewards among the target intersection and its neighbors \textit{within a city}, and the Upper Transformer learns the general decision trajectories \textit{across different cities}. This dual-level approach bolsters the model's robust generalization and transferability. Notably, when directly transferring to unseen scenarios, ours surpasses all baseline methods with \textbf{+7.91\%} on average, and even \textbf{+16.3\%} in some cases, yielding the best results.  The code is \href{https://github.com/jianghaoyuan1994/X-Light}{\textcolor{blue}{here}}.
\end{abstract}

\renewcommand{\thefootnote}{\relax}

\footnotetext{$^\dagger$Corresponding Author}
\renewcommand{\thefootnote}{\arabic{footnote}}

\section{Introduction}
An effective traffic signal control (TSC) system is the key to alleviating traffic congestion.
In recent years, Reinforcement Learning (RL) has been widely used in the field of TSC~\cite{yau2017survey}. It can interact with the environment, explore, and exploit, which helps agents discover better policies without artificial priors and assumptions. Numerous studies~\cite{2019PressLight,zheng2019learning,chen2020toward,wei2019colight,chu2019multi,lu2023dualight,du2024felight} have demonstrated its noteworthy enhancements over conventional rule-based methods~\cite{hunt1982scoot,lowrie1990scats,roess2004traffic}. 

However, most of the existing methods are \textit{scenario-specific}, meaning that the training and testing should be in the same scenario (A scenario is a simulation environment, e.g., a virtual city, with a set of intersections). When deploying on a new scenario, rebuilding the environment and re-training are needed, which involves significant costs. As a result, to the best of our knowledge, all the cities still use rule-based methods such as SCATS or SCOOT for a very fundamental reason: they can be easily reused in a new district/region/city.

This raised a critical problem that \textbf{how to orchestrate the multiple intersections for various scenarios/cities with strong generalizability.}


There are some solutions for single-agent setting: MetaLight~\cite{zang2020metalight} and GESA~\cite{jiang2023a} used one single agent to control all the intersections in a scenario and achieve transferability via gradient-based meta RL and multi-city co-training, respectively. Their drawbacks are obvious, i.e., neglecting the cooperation among the multiple intersections. Yet, learning a general Multi-Agent RL model for various scenarios is non-trivial, given various scenarios could have various road networks and traffic dynamics, rendering various environment states for each agent and collaboration patterns for multiple agents. To the best of our knowledge, only a few existing works target the same challenge: MetaVIM \cite{zhu2023metavim} and MetaGAT \cite{lou2022meta}. They both utilized the Markov Decision Processes (MDPs) trajectories 
to help the agents learn and distinguish scenario context. However, they still display limitations, such as an unstable training process and large performance drops when encountering quite dissimilar scenarios (details in Sec. \ref{sec:related}).

\begin{figure}[t]
\centering
\includegraphics[width=\columnwidth]{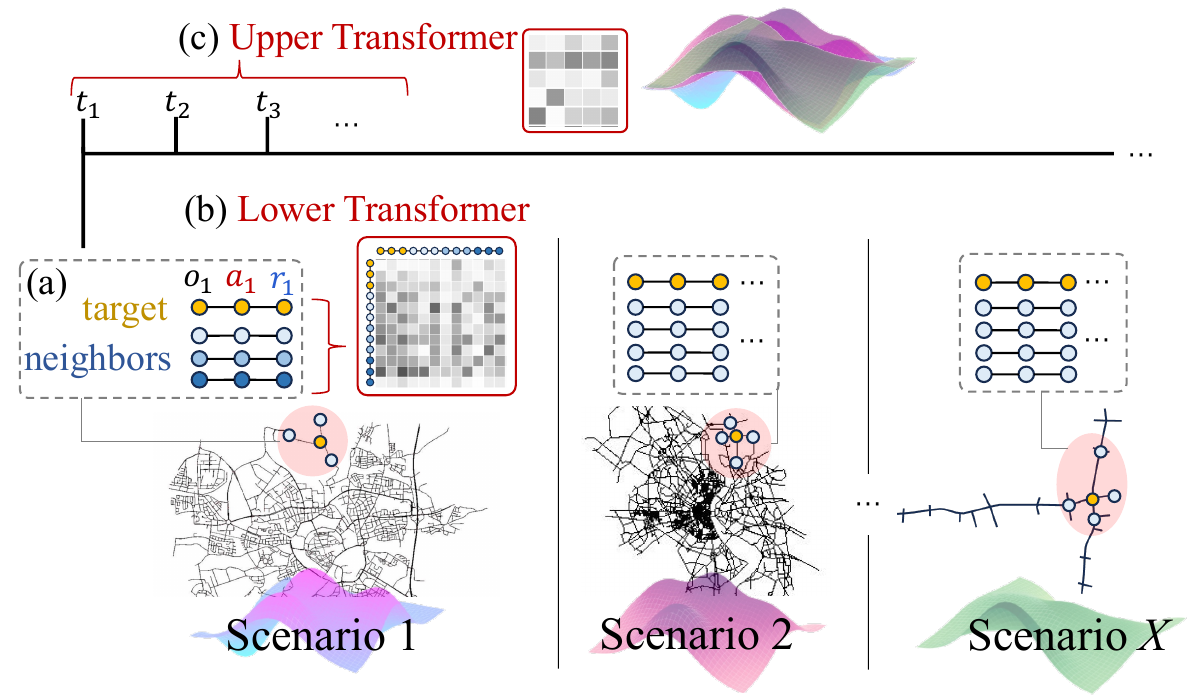} 
\caption{(a) X-Light takes the MDP $o,a,r$ trajectories of the target and its neighbors: (b) the Lower Transformer learns the attention for all the $o,a,r$-s, so that, e.g., one intersection's $o$ may have high attention with another intersection's $a$; (c) Upper Transformer learns the attention over the time through all different scenarios. 
}
\vspace{-5pt}
\label{fig: intro}
\end{figure}

To enhance cooperation and generalizability, we will incorporate the full MDP trajectories, including the observations, rewards, and actions $(o,a,r)$ of both the target intersection and its neighbors, into the method. Given the sequence nature of the trajectories from various MDPs, two natural questions are: \textbf{Q1: Can Transformer utilize the $o,a,r$ sequences of multiple intersections for better collaboration? Q2: Furthermore, can we Transformer learn the high-level cross-scenario MDPs dynamics for better transferability.} 
This forms our solution as in Figure~\ref{fig: intro}: a Transformer on Transformer (TonT) model for Meta Multi-Agent Reinforcement Learners. The \textbf{Lower Transformer} extracts single-step trajectory information from the target intersection and its neighbor intersections to encourage cooperation. 
The \textbf{Upper Transformer} learns from the historical multi-scenario multi-modal MDPs distributions and makes actions. 


The main contributions are three-fold:
\begin{itemize}
\setlength\itemsep{0.1em}
    \item In the domain of TSC, we propose the first-ever Transformer-on-Transformer (TonT) framework for meta MARL, which solves both multi-intersection collaboration and cross-city transferability/generalizability.
    \item Specifically, the \textbf{Lower Transformer} aggregates the target and its neighbors' fine-grained $o,a,r$ information and achieves better collaboration than other solutions such as the Graph Neural Network (GNN)-based methods \cite{wei2019colight,lou2022meta}, which only aggregates traffic states $o$; The \textbf{Upper Transformer}, together with a \textit{dynamic prediction} pretext task and \textit{multi-scenario co-training scheme}, learns the scenario-agnostic decision process and achieves better cross-city decision. Additionally, a \textit{residual link} is added before inputting to actor-critic for better decision.
    \item We conduct rigid experiments with various scenarios and zero-shot transfer to each, and our method is constantly the best performer. In non-transfer settings, we also achieve the most top one results.
\end{itemize}


\section{Related Work}
\label{sec:related}


\subsection{Meta Reinforcement Learning-based TSC}
Meta RL-based methods are gaining attention in TSC since they can greatly reduce adaptation costs and have promising performance in new scenarios. GeneraLight~\cite{zhang2020generalight} and AttendLight \cite{oroojlooy2020attendlight} try to generalize TSC to different traffic flow patterns: they use GAN \cite{goodfellow2014generative} and two attention models to handle traffic diversity. However, it is not strictly adapting to a new scenario. MetaLight~\cite{zang2020metalight} can be seen as the pioneering work of training a TSC agent in multiple scenarios: it uses gradient-based meta-reinforcement learning with base function reading each local data and updating local gradients, and meta-learner accumulating the memory and updating the global gradients. GESA~\cite{jiang2023a} instead proposed a general plug-in module (GPI) to make it possible to read various cities’ maps without labels and further uses large-scale scenarios to co-train an actor-critic-based agent. But all of them only considered single-agent settings and ignored the cooperation with others. However, to model the cooperative multi-agent under multiple various scenarios is quite technically challenging. Only a few models are proposed, yet their performance is quite unstable and limited.

MetaVIM~\cite{zhu2023metavim} uses context-based Meta RL to solve the generalization problem and uses intrinsic rewards to cooperate, but it is only trained in one scenario and the state does not contain neighbor information, leading to challenges in performance and adaptability when encountering an environment that is different from the training scenario. MetaGAT~\cite{lou2022meta} 
extends MetaLight with a Graph Attention Network (GAT) to aggregate the neighbors' observations into the target's and Gated Recurrent unit (GRU) to learn the MDP dynamics. However, 
it has volatile fluctuations during the training process and suboptimal performance. Potential reasons are that (1) it is trained with one scenario after another, but we mix intersections from various scenarios within one batch; (2) Relying on GAT restricts them to only utilize observation information as node feature, overlooking crucial actions and rewards within the MDP. We instead use Lower Transformer to fully construct the relation among $o,a,r$ from multiple intersections; (3) A hybrid model of combining GAT and GRU may be harder to train, and we are an elegant, Transformer-only model.

\subsection{Transformers in RL and Other Fields}
%
\textbf{Offline RL} unfolds the MDP process as a sequence of $s_t, a_t, r_t, s_{t+1}$ etc. Thus, works such as Decision Transformer (DT) \cite{chen2021decision}, PDiT \cite{mao2023pdit}, and Trajectory Transformer (TT) \cite{janner2021offline} use Transformer as the backbone, treating decision-making as a next-token prediction, i.e., to predict the next action $a_{t+1}$. TransformerLight \cite{10.1145/3580305.3599530} got inspired by DT and trained a TSC agent with offline-collected data. However, training with offline-collected data means a heavy workload to prepare high-quality data, and it cannot interact with the environment in real-time, let alone transfer to a new scenario. To increase transferability, PromptDT~\cite{xu2022prompting} proposes to add few-shot task-specific sequences as the prompt for quick adaptation for new tasks. However, offline data is still needed. 

\textbf{Transformer as Meta RL}: TrMRL~\cite{melo2022transformers} is the pioneering work that proves Transformer can function as a meta-agent. Similar to the memory reinstatement mechanism, this agent establishes connections between the immediate working memories to construct an episodic memory across the transformer layers iteratively.  Our work is the first one that applies a Transformer-based meta agent into the TSC domain; we further extend the original single-agent setting to multi-agent, with several technical challenges overcome by our final TonT solution: 
the Lower Transformer guides cooperation with neighbors utilizing $o,a,r$ information, and the upper Transformer learns meta-features across tasks, showcasing remarkable generalization capabilities.



\section{Methodology} \label{sub Method}

\begin{figure*}[t]
\centering
\includegraphics[width=\textwidth] {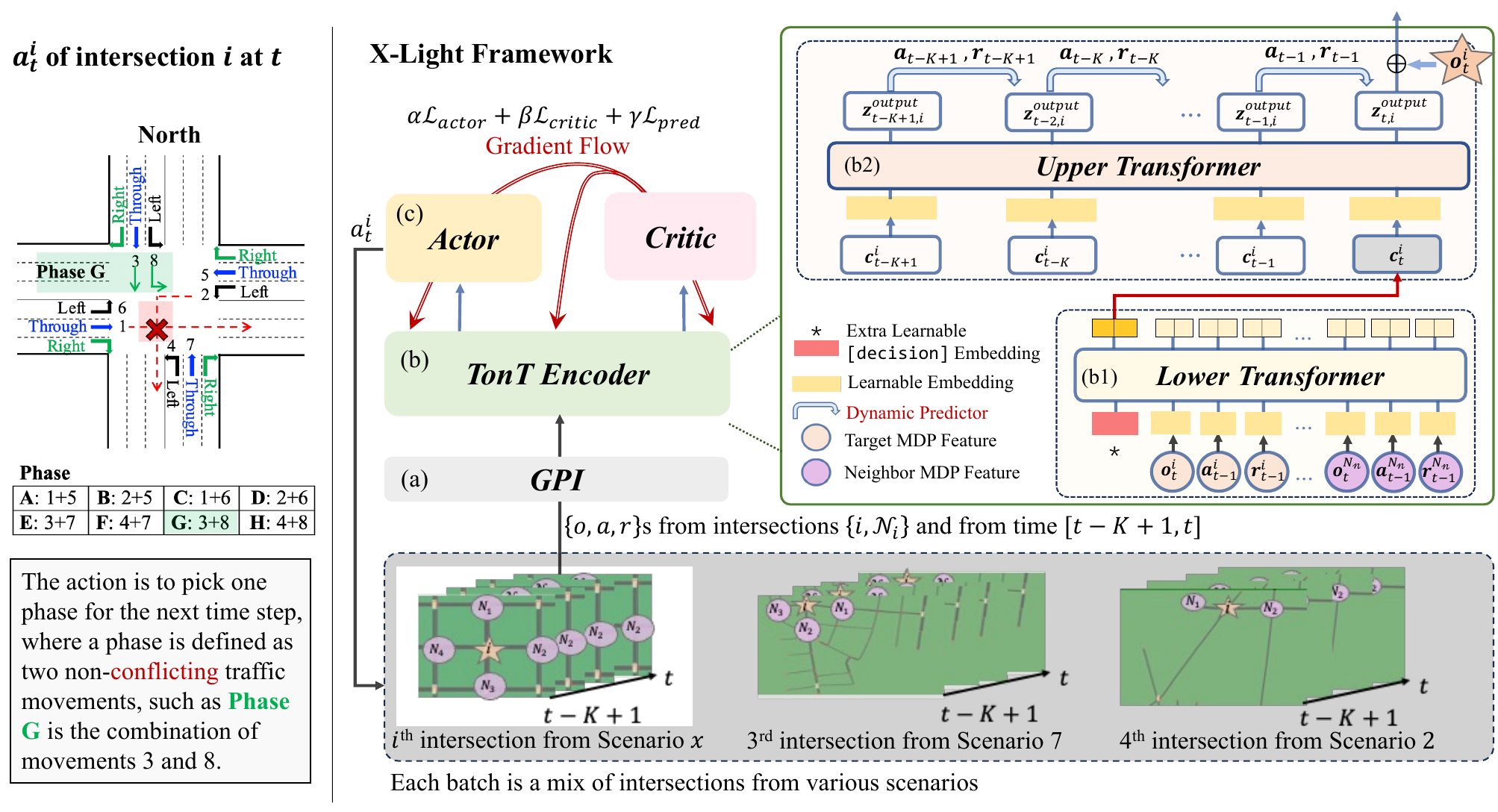} 
\vspace{-15pt}
\caption{Our method is co-trained with intersections' MDPs from various scenarios: 
(a) a GPI module unifying all the scenarios, (b) the proposed TonT Encoder, and (c) an actor-critic to make a decision. The TonT Encoder contains (b1) a Lower Transformer
aggregating the $o$, $a$, and $r$ among the target and its neighbors and (b2) an Upper Transformer learning general decisions from multi-scenario historical MDPs. 
}
\vspace{-10pt}
\label{fig: method}
\end{figure*}

This section delineates our proposed method, X-Light, a general cross-city multi-agent traffic signal control method. We begin by providing an overview of our method, followed by the introduction of the TonT Encoder module, which contains the Lower Transformer and the Upper Transformer. 
Finally, we will give detailed settings for training. The preliminary information is presented in the Appx.~\ref{sub:Preliminary}.

\subsection{Overview}\label{subsection Overview}

As shown in Fig. \ref{fig: method}, our method is trained using intersections across various scenarios within a batch. The $i$-th intersection and its neighbors $\mathcal{N}_i$ are selected, and their MDP trajectories ($o, a, r$) from time frame $[t-K+1, t]$ are sampled and fed into TonT Encoder module. It is worth noting that to allow the model to handle various intersections, we use the GPI module proposed in \cite{jiang2023a}, which maps various intersections' structure into a unified one, followed by an MLP to obtain $o_t^{\{i, \mathcal{N}_i\}}$. More details are in the Appx.~\ref{sub Implementation details}. 

As shown in Fig. \ref{fig: method}, the TonT Encoder 
employs two types of transformers: the Lower Transformer and the Upper Transformer. 
The primary role of the Lower Transformer is to integrate the target and its neighbors' MPD information at time step $t$.  It enhances agent collaboration compared with GNN-based collaboration by 6\%-13\%. The Upper Transformer utilizes historical trajectory information as context to infer the current task, thereby achieving improved transferability. 

The output of the TonT Encoder is then utilized by both the Actor and Critic to output the policy $\pi$ for executing the action and to estimate the state value for training. 

Similar to~\cite{wei2021recent}, we choose five features from the observations as states: queue length, current phase, occupancy, flow, and the number of stopping cars. The action is to choose the eight pre-defined phases for the next time interval, as shown in Fig \ref{fig: method}. At each time step $t$, the agent $i$ can choose to execute action $a_t^i$ from available action set $\mathcal{A}^i$ in the next $\Delta t$ seconds. In our experiments, we set $\Delta t$ as 15 seconds. The reward $r$ is defined as the weighted sum of queue length, wait time, delay time, and pressure. More details in Appx. \ref{sub Implementation details}.

\subsection{Lower Transformer} 
\label{subsection Lower transformer}

The Lower Transformer enhances collaboration between the target $i$ and its $n$-nearest neighbors $\mathcal{N}_i$, $\mathcal{N}_i=\{N_1,\ldots,N_n\}$. 
Existing multi-agent TSC methods only consider the states of neighboring intersections for cooperation, neglecting the complex and valuable interrelations among their observations, actions, and rewards. Considering the variations in road networks and traffic flow interconnections across diverse scenarios, one intersection's action will affect another's state and also vice versa. Thus, only relying on states is inadequate to simulate the neighbors' impacts on the target accurately. This can further result in instability within the learning process, especially during cross-scenario co-training, as we observed in MetaLight and MetaGAT's training process. 


Thus, in the Lower Transformer, we utilize the \textbf{full} MDP features of the target and its neighbors to enhance the model's understanding of the cooperation. Since we are online RL, at time $t$, we can only observe immediate $o_t$ and the previous step's $a_{t-1}, r_{t-1}$. Thus, the $i$th agent's  MDP feature at $t$ is:
\begin{equation}
\mathbf{m}_t^i = (\mathbf{o}_t^i, \mathbf{a}_{t-1}^i, \mathbf{r}_{t-1}^i)
\label{equation transition feature}
\end{equation}


Since original $\mathbf{o}, \mathbf{a}, \mathbf{r}$ have different dimensions, we first employ three trainable linear projections, i.e., $\mathbf{E}_o, \mathbf{E}_a, \mathbf{E}_r$, to map all of them to the same dimension of $d$
. Then, the full MDP transition with neighbors concatenated is defined as:
\begin{equation}
\begin{aligned}
&\mathbf{M}_t^i = [\mathbf{m}_t^i \mathbf{E};  \mathbf{m}_t^{N_1} \mathbf{E}; \ldots, \mathbf{m}_t^{N_n} \mathbf{E}] \in \mathbb{R}^{3(1+n) \times d} \\
&\text{where } \mathbf{m}_t^i \mathbf{E} = (\mathbf{o}_t^i \mathbf{E}_o, \mathbf{a}_{t-1}^i \mathbf{E}_a, \mathbf{r}_{t-1}^i \mathbf{E}_r) \in \mathbb{R}^{3 \times d}
\end{aligned}
\label{equation all transitions}
\end{equation}
If the number of neighbors is smaller than $n$, we use zero-padding and add a binary indicator embedding to all $\mathbf{o}_t$ to indicate whether this neighbor exists.

Similar to ViT \cite{dosovitskiy2020image}, a learnable \texttt{[class]} token is prepended serving as the global image representation, we also prepend a trainable \texttt{[decision]} token $\mathbf{q}_{\text{decision}} \in \mathbb{R}^d$, whose state at the output of the Lower Transformer is used as the intersection representation $\mathbf{c}_t$.
We also add standard position embedding \cite{vaswani2017attention} to each input token to retain positional information. Thus, the input to the Lower Transformer $\mathbf{z}^{\text{lower}}_{t,i}$ is:

\begin{equation}
\mathbf{z}^{\text{lower}}_{t,i} = [ \mathbf{q}_{\text{decision}}; \mathbf{m}_t^i \mathbf{E};  \mathbf{m}_t^{N_1} \mathbf{E}; \ldots, \mathbf{m}_t^{N_n} \mathbf{E} ] + \mathbf{E}^{\text{lower}}_{pos}
\label{equation Lower input2}
\end{equation}

\noindent where $\mathbf{z}^{\text{lower}}_{t,i}, \mathbf{E}_{pos}^\text{lower} \in \mathbb{R}^{(3(1+n)+1) \times d}$ and the attention in the Lower Transformer is in $\mathbb{R}^{(3(1+n)+1) \times (3(1+n)+1)}$. Then, we feed $\mathbf{z}^{\text{lower}}_{t,i}$ to the Lower transformer with 3 multi-head self-attention layers, and it outputs a capsulized intersection embedding for further decision-making, denoted as $\mathbf{c}_t^i$. 

\begin{equation}
\mathbf{c}_t^i = \text{Lower\ Transformer}(\mathbf{z}^{\text{lower}}_{t,i})
\label{equation Lower input3}
\end{equation}

\subsection{Upper Transformer} 
\label{subsection Upper transformer}
The Upper Transformer ensures that the model exhibits strong generalization in the presence of unseen intersections or scenarios. To achieve this goal, as shown in Fig. \ref{fig: method}.(b2), we employ \textit{context-based meta RL}~\cite{duan2017rl} and \textit{a dynamic predictor} within the Upper Transformer. $\mathbf{c}_{[t-K+1:t]}^i$ obtained from the Lower Transformer is first projected to dimension-$d'$ through a trainable projection denoted as $\mathbf{E}'$. Then, as the context-based meta RL, the Upper Transformer utilizes these embeddings from the last $K$ time steps. Similarly, along with the positional embedding, the input to the Upper Transformer is $ \mathbf{z}^{\text{upper}}_{[t-K+1:t],i} \in \mathbb{R}^{K \times d'}$: 
\begin{equation}
\mathbf{z}^{\text{upper}}_{[t-K+1:t],i}=[\mathbf{c}_{t-K+1}^i \mathbf{E}'; \mathbf{c}_{t-K+2}^i \mathbf{E}'; \ldots; \mathbf{c}_t^i \mathbf{E}'] + \mathbf{E}^{\text{upper}}_{pos},
\label{equation Upper input1}
\end{equation} 
The attention mechanism in the Upper Transformer (consisting of 3 multi-head self-attention layers) operates within the $\mathbb{R}^{K \times K}$ dimension, capturing the environmental dynamics related to the target intersection from historical features.
\begin{equation}
\mathbf{z}^{\text{output}}_{[t-K+1:t],i} = \text{Upper\ Transformer}(\mathbf{z}^{\text{upper}}_{[t-K+1:t],i})
\label{equation Upper input3}
\end{equation}

\textbf{Dynamic Predictor}: To enhance the agent's comprehension of the current task and the influence between intersections, we introduce a dynamic predictor between the Upper Transformer's each time-step output, enhancing the agent to learn the environment dynamics. The dynamic predictor is a pretext prediction task to conduct autoregression prediction, encouraging the Upper Transformer to capture the cross-scenario dynamics. As shown in Fig. \ref{fig:ablation}, this dynamic predictor can improve performance by \textbf{4.4\%}. Specifically, we feed the previous time step $\mathbf{z}^{\text{output}}_{t-1, i}$, concatenated with all the actions $\mathbf{a}_{t-1}^{\{i, \mathcal{N}_i\}}$ and rewards $\mathbf{r}_{t-1}^{\{i, \mathcal{N}_i\}}$ from the target and its neighbors, to predict the next  $\mathbf{z}^{\text{output}}_{t, i}$:
\begin{equation}
\begin{aligned}
\label{equation Upper input4}
\hat{\mathbf{z}}^{\text{output}}_{t, i} =  \text{MLP}([\mathbf{z}^{\text{output}}_{t-1, i}; \mathbf{a}_{t-1}^{\{i, \mathcal{N}_i\}}; \mathbf{r}_{t-1}^{\{i, \mathcal{N}_i\}}])
\end{aligned}
\end{equation}

A prediction loss based on mean squared error (MSE) is:
\begin{equation}
\mathcal{L}_{pred} = \text{MSE}(\hat{\mathbf{z}}^{\text{output}}_{t, i}, \mathbf{z}^{\text{output}}_{t, i})
\label{equation predictor loss}
\end{equation}

\subsection{Actor-Critic}

For the decision policy $\pi$, we use the PPO method~\cite{schulman2017proximal}, both Actor and Critic are two-layer MLPs. The actor receives the TonT Encoder's output and makes the action $a_i^t$ for the target intersection $i$.

\textbf{Residual Link}: However, before inputting into the actor-critic module, to avoid over-abstracting the intersection's embedding through the TonT Encoder, we directly add the observation $\mathbf{o}_t^i$ of the target intersection $i$ into the Upper Transformer's output $\mathbf{z}^{\text{output}}_{t,i}$. 
This ensures the embedding input into the actor has enough observational information from the target intersection. 
\begin{equation}
\mathbf{a}_t^i \sim \pi (\cdot | \mathbf{z}^{\text{output}}_{t,i} + \mathbf{o}_t^i)    
\label{equation Upper input5}
\end{equation}

The $\mathbf{z}^{\text{output}}_{t,i}$ is the embedding after TonT Encoder's feature abstraction, which is particularly essential for scenarios with complex intercorrelations; $\mathbf{o}_t^i$ instead is the direct self-observation, which is proven as rather essential for simple scenarios, where focusing on yourself is enough. As shown in Fig.\ref{fig:ablation}, without residual link can lead to a \textbf{2\%} performance drop. Thus, our design strikes a good balance for scenarios in various complexity.
Therefore, together with Eq. (\ref{equation predictor loss}), the overall optimization objective can be formatted as:
\begin{equation}
\mathcal{L} = \alpha \mathcal{L}_{actor} + \beta \mathcal{L}_{critic} + \gamma \mathcal{L}_{pred} , 
\label{equation all loss}
\end{equation}
where $\alpha, \beta, \gamma$ are tuning parameters. The Actor loss and Critic loss are the same as PPO. 

\subsection{Multi-scenario Co-Training}
\label{subsection Detailed training settings}

To further increase model generality, we employed the multi-scenario co-train to increase data diversity. Unlike MetaGAT~\cite{lou2022meta} or MetaLight~\cite{zang2020metalight}, which utilizes the multi-scenario sequential training, i.e., within a batch, intersections are from the same scenario, and scenarios are read until previous one is all seen, we adopted multi-scenario co-training, i.e., within each batch, intersections are stochastically chosen from various scenarios. Given each intersection from each scenario has a distinct structure, we adopt a general preprocess module, GPI from GESA~\cite{jiang2023a}, whose core idea is to map all various-structured intersections into a unified 4-leg one. This encourages the agent to learn generalist knowledge and also 
enables a more stable training process.


Lastly, the learning algorithm is shown in Algorithm~\ref{alg:algo}.

\begin{algorithm}[t]
\caption{X-Light training process}
\label{alg:algo}
\textbf{Input}: A set of target intersections $\mathcal{I}$ from a set of multi-agent scenarios $\mathcal{X}$; training episodes \textit{E}; the number of neighbor $n$; the Upper Transformer input length $K$; 

\textbf{Initialize}: buffer $\mathcal{D}$; parameters $\theta$;

\textbf{Output}: Optimized policy $\pi^{\theta}$
\begin{algorithmic}[1]
\FOR {episode=1, $\ldots, \textit{E}$}
\STATE clean buffer $\mathcal{D}\leftarrow \emptyset$;
\FOR {each scenario $x$}
\STATE Find nearest $n$ neighbors $\mathcal{N}$ of each intersection;
\FOR {each time step $t$}
\STATE Get the last $K$ transitions $\{\mathbf{m}_{t-k+1}^i\}_{k=1,\ldots, K}^x$ of each intersection $i$ according to Eq. (1) and add these into  $\mathcal{D}$;
\STATE Get action $\mathbf{a}_t^i$ according to Eq. (\ref{equation Upper input5}) and take joint action $\{\mathbf{a}_t^1,...,\mathbf{a}_t^{n_x}\}$;
\STATE Get the the next states $\mathbf{o}^i_{t+1}$, and rewards $\mathbf{r}_t^i$;
\ENDFOR
\ENDFOR
\FOR{each step in training steps}
\STATE sample minibatch data from $\mathcal{D}$;
\STATE Get dynamic predictions $\{\hat{\mathbf{z}}^{\text{output}}_{t-k+1, i}\}_{k=1,\ldots, K-1}$ according to Eq.(\ref{equation Upper input4});
\STATE Computer $\mathcal{L}$ according (\ref{equation all loss}) and update parameter $\theta$;
\ENDFOR
\ENDFOR
\end{algorithmic}
\end{algorithm}


\section{Experiments} \label{sub Experiments}

\subsection{Datasets}
We conducted experiments on the simulation of urban mobility (SUMO)\footnote{https://www.eclipse.org/sumo/} as the simulator. The duration of each episode is 3600 seconds. 
Seven different scenarios~\cite{ault2021reinforcement,jiang2023a} are employed in our co-training: five of them are reported in results, and another two scenarios (\textit{Fenglin} and \textit{Nanshan}) are only employed to increase the number of meta-training scenarios. To enable multi-scenario co-training, we utilize the GPI module in GESA to significantly reduce the need for manual labeling and unify the observation space and action space of all intersections. The main idea of GPI is to map all variously structured intersections into a unified 4-leg one using the relative angle and masking. For details, please refer to ~\cite{jiang2023a}. 
Table \ref{tab: Data statistics of datasets1} shows the properties of different scenarios. The details about these scenarios are in Appx.~\ref{sub Detail statistics}.

\begin{table}[t]
\centering
\small
\resizebox{0.99\columnwidth}{!}{%
\begin{tabular}{c|lll||l|lll}
\toprule
\textbf{Scenarios}      & Country & Type & \#Total Int. & \textbf{Scenarios}      & Country & Type & \#Total Int. \\ \hline
\textit{\textbf{Grid $4 \times 4$}}   & synthetic & region & 16  &  \textit{\textbf{Cologne8}}     & Germany & region & 8                           \\
\textit{\textbf{Avenue $4 \times 4$}}   & synthetic & region & 16  &  \textit{\textbf{Ingolstadt21}} & Germany & region & 12                   \\
\textit{\textbf{Grid $5 \times 5$}}     & synthetic & region & 25  & \textit{\textbf{Fenglin}}     & China & corridor & 7                              \\                           
     &  &  &   & \textit{\textbf{Nanshan}} & China & region & 28                             \\ \bottomrule
\end{tabular}
}
\caption{Statistics of the scenarios.}
\label{tab: Data statistics of datasets1}
\vspace{-5pt}
\end{table}

\begin{table*}[t] \small
\resizebox{\textwidth}{!}{%
\begin{tabular}{c|lllll|lllll}
\toprule
\multirow{2}{*}{\textbf{Methods}} & \multicolumn{5}{c|}{\textbf{Avg. Trip Time (seconds)}}                                                                                                                                                                 & \multicolumn{5}{c}{\textbf{Avg. Delay Time (seconds)}}                                                                                                                                      \\ \cline{2-11} 
                         & \multicolumn{1}{c}{\textit{transfer to  Grid4$\times$4}} & \multicolumn{1}{c}{\textit{transfer to  Grid5$\times$5}} & \multicolumn{1}{c}{\textit{transfer to  Arterial4$\times$4}} & \multicolumn{1}{c}{\textit{transfer to  Ingolstadt21}} & \multicolumn{1}{c|}{\textit{transfer to Cologne8}} & \multicolumn{1}{c}{\textit{transfer to Grid4$\times$4}} & \multicolumn{1}{c}{  \textit{ transfer to Grid5$\times$5}} & \multicolumn{1}{c}{\textit{transfer to Arterial4$\times$4}} & \multicolumn{1}{c}{\textit{transfer to Ingolstadt21}} & \multicolumn{1}{c}{\textit{transfer to Cologne8}} \\ \hline
FTC                      & 206.68 \fontsize{8pt}{8pt}\selectfont{± 0.54}                                                & 550.38 \fontsize{8pt}{8pt}\selectfont{± 8.31}                      & 828.38 \fontsize{8pt}{8pt}\selectfont{± 8.17}                          & 319.41 \fontsize{8pt}{8pt}\selectfont{± 24.48}                   & 124.4 \fontsize{8pt}{8pt}\selectfont{± 1.99}                  & 94.64 \fontsize{8pt}{8pt}\selectfont{± 0.43}                       & 790.18 \fontsize{8pt}{8pt}\selectfont{± 7.96}                      & 1234.30 \fontsize{8pt}{8pt}\selectfont{± 6.50}                         & 183.70 \fontsize{8pt}{8pt}\selectfont{± 26.21}                   & 62.38 \fontsize{8pt}{8pt}\selectfont{± 2.95}                 \\
MP              & 175.97 \fontsize{8pt}{8pt}\selectfont{± 0.70}                                                & 274.15 \fontsize{8pt}{8pt}\selectfont{± 15.23}                     & 686.12 \fontsize{8pt}{8pt}\selectfont{± 9.57}                          & 375.25 \fontsize{8pt}{8pt}\selectfont{± 2.40}                    & 95.96 \fontsize{8pt}{8pt}\selectfont{± 1.11}                  & 64.01 \fontsize{8pt}{8pt}\selectfont{± 0.71}                       & 240.00 \fontsize{8pt}{8pt}\selectfont{± 18.43}                     & 952.53 \fontsize{8pt}{8pt}\selectfont{± 12.48}                         & 275.36 \fontsize{8pt}{8pt}\selectfont{± 14.38}                   & 31.93 \fontsize{8pt}{8pt}\selectfont{± 1.07}                 \\ \hline
MetaLight                & 169.21 \fontsize{8pt}{8pt}\selectfont{± 1.16}                                                & {\ul 244.99 \fontsize{8pt}{8pt}\selectfont{± 8.18}}                      & 392.34 \fontsize{8pt}{8pt}\selectfont{± 4.39}                          & 298.67 \fontsize{8pt}{8pt}\selectfont{± 5.09}                    & 92.38 \fontsize{8pt}{8pt}\selectfont{± 0.94}                  & 57.82 \fontsize{8pt}{8pt}\selectfont{± 0.67}                       & 202.32\fontsize{8pt}{8pt}\selectfont{ ± 11.20}                     & 850.42 \fontsize{8pt}{8pt}\selectfont{± 36.35}                         & {\ul 166.35 \fontsize{8pt}{8pt}\selectfont{± 5.53}}                    & 28.37 \fontsize{8pt}{8pt}\selectfont{± 0.73}                 \\
GESA                     & 166.23 \fontsize{8pt}{8pt}\selectfont{± 1.07}                                                &   284.05 \fontsize{8pt}{8pt}\selectfont{± 25.36}                                 & 410.59 \fontsize{8pt}{8pt}\selectfont{± 2.60}                          & 318.30 \fontsize{8pt}{8pt}\selectfont{± 9.56}                    & {\ul 88.76 \fontsize{8pt}{8pt}\selectfont{± 0.46}}                  & 54.69 \fontsize{8pt}{8pt}\selectfont{± 0.81}                       &  246.96 \fontsize{8pt}{8pt}\selectfont{± 37.62}                                  & 972.87 \fontsize{8pt}{8pt}\selectfont{± 114.04}                        & 208.41 \fontsize{8pt}{8pt}\selectfont{± 12.01}                   & {\ul 25.13 \fontsize{8pt}{8pt}\selectfont{± 0.64}}                 \\ \hline
MetaGAT                  & 165.18 \fontsize{8pt}{8pt}\selectfont{± 0.00}                                                & 278.67 \fontsize{8pt}{8pt}\selectfont{± 0.00}                      &379.47 \fontsize{8pt}{8pt}\selectfont{± 0.00}                          & {\ul 288.10 \fontsize{8pt}{8pt}\selectfont{± 0.00}}           & 90.41 \fontsize{8pt}{8pt}\selectfont{± 0.00}                  & 52.71 \fontsize{8pt}{8pt}\selectfont{± 0.00}                       & 241.51 \fontsize{8pt}{8pt}\selectfont{± 0.00}                      & {\ul 753.88 \fontsize{8pt}{8pt}\selectfont{± 0.00}}                          & 177.10 \fontsize{8pt}{8pt}\selectfont{± 0.00}                     & 26.59 \fontsize{8pt}{8pt}\selectfont{± 0.00}                 \\ \cline{2-11}
Ours$_{GNN}$                 & {\ul 164.46 \fontsize{8pt}{8pt}\selectfont{± 0.00}}                                                & 249.64 \fontsize{8pt}{8pt}\selectfont{± 0.00}                     & {\ul 366.97 \fontsize{8pt}{8pt}\selectfont{± 0.00}}                 & 298.97 \fontsize{8pt}{8pt}\selectfont{± 0.00}                    & 91.02 \fontsize{8pt}{8pt}\selectfont{± 0.00}                  & {\ul 52.43 \fontsize{8pt}{8pt}\selectfont{± 0.00}}                       & {\ul 194.36 \fontsize{8pt}{8pt}\selectfont{± 0.00}}                      & 762.61 \fontsize{8pt}{8pt}\selectfont{± 0.00}                 & 178.69 \fontsize{8pt}{8pt}\selectfont{± 0.00}                    & 27.26 \fontsize{8pt}{8pt}\selectfont{± 0.00}                 \\ 
Ours                     & \textbf{162.65 \fontsize{8pt}{8pt}\selectfont{± 0.00}} (+1.5\%)                                       & \textbf{243.26 \fontsize{8pt}{8pt}\selectfont{± 6.49}}  (+12.9\%)           & \textbf{361.38 \fontsize{8pt}{8pt}\selectfont{± 0.00}}    (+5.1\%)                      & \textbf{280.20 \fontsize{8pt}{8pt}\selectfont{± 0.00}} (+2.7\%)                   & \textbf{88.49 \fontsize{8pt}{8pt}\selectfont{± 0.00}} (+2.1\%)        & \textbf{50.90 \fontsize{8pt}{8pt}\selectfont{± 0.89}} (+3.4\%)             & \textbf{193.84 \fontsize{8pt}{8pt}\selectfont{± 0.00}}    (+19.7\%)         & \textbf{705.48 \fontsize{8pt}{8pt}\selectfont{± 0.00}}    (+6.4\%)                      & \textbf{160.42 \fontsize{8pt}{8pt}\selectfont{± 0.00}}   (+9.6\%)        & \textbf{24.71 \fontsize{8pt}{8pt}\selectfont{± 0.00}}  (+7.1\%)      \\  \bottomrule
\end{tabular}}
\vspace{-5pt}
\caption{Performance when transferring to an unseen scenario, with the format as ``mean ± standard deviation (gain in \% compared with the best baseline)'', the best \textbf{boldfaced} and second best \underline{underlined}. We employ zero-shot transfer for evaluation: for each test scenario, we employ six other scenarios during the training process and then directly evaluate them in the respective test scenario. Our proposed method achieves the best performance in zero-shot transfer. (Some standard deviation = 0.00 because only two digits are kept.)}
\label{table:unseen-table}
\vspace{-5pt}
\end{table*}

\begin{table*}[t] 
\centering 
\resizebox{\textwidth}{!}{%
\begin{tabular}{c|lllll|lllll}
\toprule
\multirow{2}{*}{\textbf{Methods}} & \multicolumn{5}{c|}{\textbf{Avg. Trip Time (seconds)}}                                                                                                           & \multicolumn{5}{c}{\textbf{Avg. Delay Time (seconds)}}                                                       \\ \cline{2-11} 

                                  & \multicolumn{1}{c}{\textit{Grid4$\times$4 (seen)}} & \multicolumn{1}{c}{\textit{Grid5$\times$5 (seen)}} & \multicolumn{1}{c}{\textit{Arterial4$\times$4 (seen)}} & \multicolumn{1}{c}{\textit{Ingolstadt21 (seen)}} & \multicolumn{1}{c|}{\textit{Cologne8 (seen)}} & \multicolumn{1}{c}{\textit{Grid4$\times$4 (seen)}} & \multicolumn{1}{c}{\textit{Grid5$\times$5 (seen)}} & \multicolumn{1}{c}{\textit{Arterial4$\times$4 (seen)}} & \multicolumn{1}{c}{\textit{Ingolstadt21 (seen)}} & \multicolumn{1}{c}{\textit{Cologne8 (seen)}} \\ \hline

FTC                               & 206.68 \fontsize{8pt}{8pt}\selectfont{± 0.54}               & 550.38 \fontsize{8pt}{8pt}\selectfont{± 8.31}               & 828.38 \fontsize{8pt}{8pt}\selectfont{± 8.17}                   & 319.41 \fontsize{8pt}{8pt}\selectfont{± 24.48}                   & 124.4 \fontsize{8pt}{8pt}\selectfont{± 1.99}                  & 94.64 \fontsize{8pt}{8pt}\selectfont{± 0.43}                & 790.18 \fontsize{8pt}{8pt}\selectfont{± 7.96}               & 1234.30 \fontsize{8pt}{8pt}\selectfont{± 6.50}                  & 183.70 \fontsize{8pt}{8pt}\selectfont{± 26.21}                   & 62.38 \fontsize{8pt}{8pt}\selectfont{± 2.95}                 \\

MaxPressure                       & 175.97 \fontsize{8pt}{8pt}\selectfont{± 0.70}               & 274.15 \fontsize{8pt}{8pt}\selectfont{± 15.23}              & 686.12 \fontsize{8pt}{8pt}\selectfont{± 9.57}                   & 375.25 \fontsize{8pt}{8pt}\selectfont{± 2.40}                    & 95.96 \fontsize{8pt}{8pt}\selectfont{± 1.11}                  & 64.01 \fontsize{8pt}{8pt}\selectfont{± 0.71}                & 240.00 \fontsize{8pt}{8pt}\selectfont{± 18.43}              & 952.53 \fontsize{8pt}{8pt}\selectfont{± 12.48}                  & 275.36 \fontsize{8pt}{8pt}\selectfont{± 14.38}                   & 31.93 \fontsize{8pt}{8pt}\selectfont{± 1.07}      \\ \hline

MPLight                           & 179.51 \fontsize{8pt}{8pt}\selectfont{± 0.95}                 & 261.76 \fontsize{8pt}{8pt}\selectfont{± 6.60}               & 541.29 \fontsize{8pt}{8pt}\selectfont{± 45.24}                  & 319.28 \fontsize{8pt}{8pt}\selectfont{± 10.48}                   & 98.44 \fontsize{8pt}{8pt}\selectfont{± 0.62}                  & 67.52 \fontsize{8pt}{8pt}\selectfont{± 0.97}                & 213.78 \fontsize{8pt}{8pt}\selectfont{± 14.44}              & 1083.18 \fontsize{8pt}{8pt}\selectfont{± 63.38}                 & 185.04 \fontsize{8pt}{8pt}\selectfont{± 10.70}                   & 34.38 \fontsize{8pt}{8pt}\selectfont{± 0.63}                 \\    
IPPO                              & 167.62 \fontsize{8pt}{8pt}\selectfont{± 2.42}               & 259.28 \fontsize{8pt}{8pt}\selectfont{± 9.55}               & 431.31 \fontsize{8pt}{8pt}\selectfont{± 28.55}                  & 379.22 \fontsize{8pt}{8pt}\selectfont{± 34.03}                   & 90.87 \fontsize{8pt}{8pt}\selectfont{± 0.40}                  & 56.38 \fontsize{8pt}{8pt}\selectfont{± 1.46}                & 243.58 \fontsize{8pt}{8pt}\selectfont{± 9.29}               & 914.58 \fontsize{8pt}{8pt}\selectfont{± 36.90}                  & 247.68 \fontsize{8pt}{8pt}\selectfont{± 35.33}                   & 26.82 \fontsize{8pt}{8pt}\selectfont{± 0.43}    \\          

rMAPPO                            & 164.96 \fontsize{8pt}{8pt}\selectfont{± 1.87}               & 300.90 \fontsize{8pt}{8pt}\selectfont{± 8.31}               & 565.67 \fontsize{8pt}{8pt}\selectfont{± 44.8}                   & 453.61 \fontsize{8pt}{8pt}\selectfont{± 29.66}                   & 97.68 \fontsize{8pt}{8pt}\selectfont{± 2.03}                  & 53.65 \fontsize{8pt}{8pt}\selectfont{± 1.00}                & 346.78 \fontsize{8pt}{8pt}\selectfont{± 28.25}              & 1185.2 \fontsize{8pt}{8pt}\selectfont{± 167.48}                 & 372.2 \fontsize{8pt}{8pt}\selectfont{± 39.85}                    & 33.37 \fontsize{8pt}{8pt}\selectfont{± 1.97}    \\             
CoLight                           & 163.52 \fontsize{8pt}{8pt}\selectfont{± 0.00}               & 242.37 \fontsize{8pt}{8pt}\selectfont{± 0.00}               & 409.93 \fontsize{8pt}{8pt}\selectfont{± 0.00}                   & 337.46 \fontsize{8pt}{8pt}\selectfont{± 0.00}                    & 89.72 \fontsize{8pt}{8pt}\selectfont{± 0.00}                  & 51.58 \fontsize{8pt}{8pt}\selectfont{± 0.00}                & 248.32 \fontsize{8pt}{8pt}\selectfont{± 0.00}               & 776.61 \fontsize{8pt}{8pt}\selectfont{± 0.00}                   & 226.06 \fontsize{8pt}{8pt}\selectfont{± 0.00}                    & 25.56 \fontsize{8pt}{8pt}\selectfont{± 0.00}  \\ \hline

MetaLight                         & 169.21 \fontsize{8pt}{8pt}\selectfont{± 1.26}               & 247.83 \fontsize{8pt}{8pt}\selectfont{± 5.99}              & 381.77 \fontsize{8pt}{8pt}\selectfont{± 12.85}                  & 292.26 \fontsize{8pt}{8pt}\selectfont{± 4.40}                    & 91.57 \fontsize{8pt}{8pt}\selectfont{± 0.75}                  & 57.56 \fontsize{8pt}{8pt}\selectfont{± 0.76}                & 209.13 \fontsize{8pt}{8pt}\selectfont{± 19.40}             & 862.32 \fontsize{8pt}{8pt}\selectfont{± 39.01}                  & {\ul 164.80 \fontsize{8pt}{8pt}\selectfont{± 3.75}}                    & 27.61 \fontsize{8pt}{8pt}\selectfont{± 0.78}                 \\

GESA                              & \textbf{161.33 \fontsize{8pt}{8pt}\selectfont{± 1.34}}      & 252.11 \fontsize{8pt}{8pt}\selectfont{± 9.94}               & 393.57 \fontsize{8pt}{8pt}\selectfont{± 13.72}                  & 320.02 \fontsize{8pt}{8pt}\selectfont{± 5.57}                   & 90.59 \fontsize{8pt}{8pt}\selectfont{± 0.74}                  & \textbf{49.60 \fontsize{8pt}{8pt}\selectfont{± 0.71}}       & 210.74 \fontsize{8pt}{8pt}\selectfont{± 13.56}              & 775.22 \fontsize{8pt}{8pt}\selectfont{± 8.63}                   & 209.57 \fontsize{8pt}{8pt}\selectfont{± 3.32}                    & 26.50 \fontsize{8pt}{8pt}\selectfont{± 0.87}      \\ \hline

MetaGAT                           & 165.23 \fontsize{8pt}{8pt}\selectfont{± 0.00}                            & 266.60 \fontsize{8pt}{8pt}\selectfont{± 0.00}                             & {\ul 374.80 \fontsize{8pt}{8pt}\selectfont{± 0.87}}                                & {\ul 290.73 \fontsize{8pt}{8pt}\selectfont{± 0.45}}                                & 90.74 \fontsize{8pt}{8pt}\selectfont{± 0.00}                              & 53.20 \fontsize{8pt}{8pt}\selectfont{± 0.00}                            &  234.80 \fontsize{8pt}{8pt}\selectfont{± 0.00}                           & 772.36 \fontsize{8pt}{8pt}\selectfont{± 0.00}                              & 176.86 \fontsize{8pt}{8pt}\selectfont{± 2.37}                                 &  26.85 \fontsize{8pt}{8pt}\selectfont{± 0.00}                            \\ \cline{2-11}
Ours$_{GNN}$                          & 164.32 \fontsize{8pt}{8pt}\selectfont{± 0.00}         &  {\ul 233.12 \fontsize{8pt}{8pt}\selectfont{± 0.00}}     & 382.38 \fontsize{8pt}{8pt}\selectfont{± 0.00}      & 319.98 \fontsize{8pt}{8pt}\selectfont{± 0.00}     & {\ul 89.29 \fontsize{8pt}{8pt}\selectfont{± 0.00}}      &   51.83 \fontsize{8pt}{8pt}\selectfont{± 0.00}       &   {\ul 204.24 \fontsize{8pt}{8pt}\selectfont{± 0.00}}    & {\ul 710.68 \fontsize{8pt}{8pt}\selectfont{± 0.00}}        &  184.10 \fontsize{8pt}{8pt}\selectfont{± 0.00}          & {\ul 25.00 \fontsize{8pt}{8pt}\selectfont{± 0.00}} \\
Ours                              & {\ul 162.47 \fontsize{8pt}{8pt}\selectfont{± 0.00}}    (-0.7\%)            & \textbf{220.63 \fontsize{8pt}{8pt}\selectfont{± 0.00}}    (+9.0\%)   & \textbf{349.60\fontsize{8pt}{8pt}\selectfont{± 0.00}}    (+6.7\%)       & \textbf{278.05 \fontsize{8pt}{8pt}\selectfont{± 0.00}}    (+4.4\%)        & \textbf{88.55 \fontsize{8pt}{8pt}\selectfont{± 0.00}}   (+1.4\%)       & {\ul 50.27 \fontsize{8pt}{8pt}\selectfont{± 0.00}}             (-1.3\%)    & \textbf{187.74 \fontsize{8pt}{8pt}\selectfont{± 0.00}}  (+11.0\%)     & \textbf{697.79\fontsize{8pt}{8pt} \selectfont{± 0.00}}     (+9.7\%)      & \textbf{160.39 \fontsize{8pt}{8pt}\selectfont{± 0.00}}   (+2.7\%)         & \textbf{24.31 \fontsize{8pt}{8pt}\selectfont{± 0.00}}   (+4.9\%)      \\ \bottomrule
\end{tabular}}
\vspace{-5pt}
\caption{Performance on various scenarios that is seen in the training: methods from FTC to CoLight are trained and tested on the same scenario of each column, and methods from MetaLight to Ours are trained with all seven scenarios, five of them have results in each column.
} 
\label{table:main-table}
\vspace{-10pt}
\end{table*}

\subsection{Baselines}
To comprehensively verify the effectiveness of our proposed method, we employed four types of methods for a comprehensive comparison: Conventional methods, Multi-agent methods, Meta-Learning methods, and a combination of both Meta-Learning and Multi-agent methods. \\
\textbf{Conventional methods}:
\begin{itemize}
\setlength\itemsep{0.1em}
    \item \textbf{Fixed Time Control (FTC)}~\cite{roess2004traffic}  with random offset executes each phase within a loop, utilizing a pre-defined phase duration.
    \item \textbf{MaxPressure}~\cite{kouvelas2014maximum} is a powerful conventional method that consistently selects the phase with the highest pressure among all phases.  
\end{itemize}
\textbf{Multi-agent methods}:
\begin{itemize}
\setlength\itemsep{0.1em}
    \item \textbf{MPLight}~\cite{chen2020toward} is based on a phase competition mechanism to select which phase to execute, and utilizes the concept of pressure as state and reward to coordinate multiple intersections.
    \item \textbf{IPPO}~\cite{ault2021reinforcement} controls each intersection with an independent PPO agent, and the agents at each intersection have the same model architecture but different model parameters. 
    \item \textbf{rMAPPO}~\cite{yu2022surprising} is executed with independent PPO agents like IPPO, but it uses the overall information of all intersections to jointly optimize traffic efficiency. Furthermore, we use RNNs to introduce historical information into the agent.
    \item \textbf{CoLight}~\cite{wei2019colight} utilizes the GAT to aggregate information from neighboring intersections, enhancing cooperation between intersections.
\end{itemize}
\textbf{Meta-Learning and Single-agent methods}:
\begin{itemize}
\setlength\itemsep{0.1em}
    \item \textbf{MetaLight}~\cite{zang2020metalight} using the meta-learning method to train multiple scenarios to increase the generalization of the model.
    \item \textbf{GESA}~\cite{jiang2023a} proposes a unified state and action space, and subsequently employs the GPI module for large-scale multi-scenario collaborative training, which improves performance and generality.
\end{itemize}
\textbf{Meta-Learning and Multi-agent methods}:
\begin{itemize}
\setlength\itemsep{0.1em}
    \item \textbf{MetaGAT}~\cite{lou2022meta} combines contextual meta-learning based on GRU to improve generalization and GAT for cooperation. Yet, the GRU is not scenario-agnostic trained, and GAT only uses observations $o$.
    \item \textbf{X-Light} (ours) and \textbf{X-Light$_{\text{GNN}}$}, which replaces the Lower Transformer with GNN~\cite{hamilton2017inductive}, and the node feature of each intersection is obtained by concatenating $o,a,r$ into a three times longer embedding, discarding the interrelations among them. 
\end{itemize}

\subsection{Evaluation Metrics}
As same as most existing works~\cite{lou2022meta,oroojlooy2020attendlight,chen2020toward}, we use \textbf{average trip time} as a component of evaluation metrics, defined as the average time for each vehicle from entering the scenario to leaving the scenario. However, merely using the average trip time cannot accurately reflect the real traffic situation, e.g., in the cases of severe traffic congestion, new vehicles are prevented from entering the scenario, which leads to a relatively low average trip time though it is a severe traffic condition. Thus, we also add \textbf{average delay time}~\cite{ault2021reinforcement} for evaluation, defined as the delay caused by signalized intersections and traffic congestion. 


\subsection{Results}
In this section, we show our proposed model's superior transferability and general great performance. 

\begin{figure}[t]
\centering
\includegraphics[width=0.95\columnwidth]{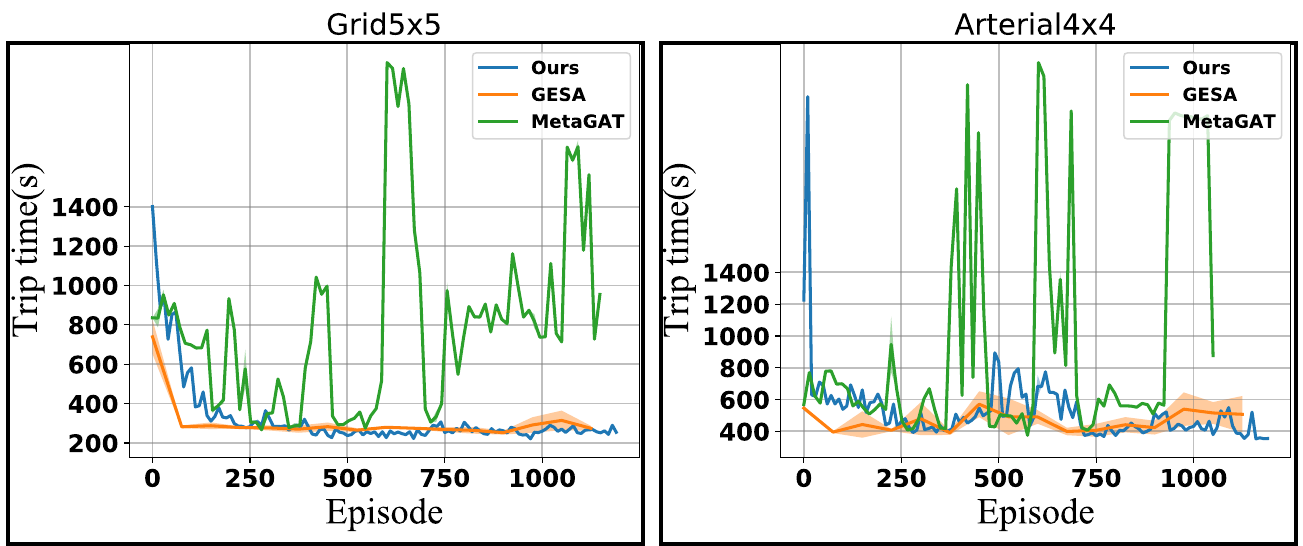} 
\caption{Trip time during the training process of Top 3 methods.}
\label{fig:duration_compare}
\vspace{-10pt}
\end{figure}

\subsubsection{Great Transferability When Handling A New Scenario}
We use a zero-shot way to evaluate each model's performance in transferring to new scenarios. Only transferable models are selected. In Table \ref{table:unseen-table}, each column means that we use the other six scenarios during training and then directly transfer to this unseen scenario. 
Our model achieves the best transfer results in all scenarios. \textbf{ (1) Cooperation is needed:} Compared with single agent methods, e.g., MetaLight, cooperation is needed for better transferability. \textbf{ (2) TonT is better for Meta MARL:} compared to the second-best MetaGAT, ours achieved a \textbf{+7.91\%} improvement on average and \textbf{+16.3\%} in \textit{Grid5$\times$5}, mainly because our unified Transformer on Transformer design captures collaborators' $o,a,r$ interdependency for better local cooperation, and global cross-scenario dynamics through multi-scenario co-training, respectively. Yet, MetaGAT only focuses on the neighbors' $o$ interdependency and lacks a scenario-agnostic training scheme. \textit{In some cases (\textit{Grid5$\times$5, Ingostadt21}), MetaGAT cannot beat MetaLight, which means that when cooperation is not as well designed as ours, it can even worsen the transferability when dealing with multi-scenarios.} \textbf{ (3) Lower Transformer is better for cooperation:} ours$_{GNN}$ flattens $o,a,r$, thus losing the $o,a,r$ interdependency, further highlighting the necessity of the Lower Transformer. 

\begin{figure}[t]
\centering
\includegraphics[width=\columnwidth]{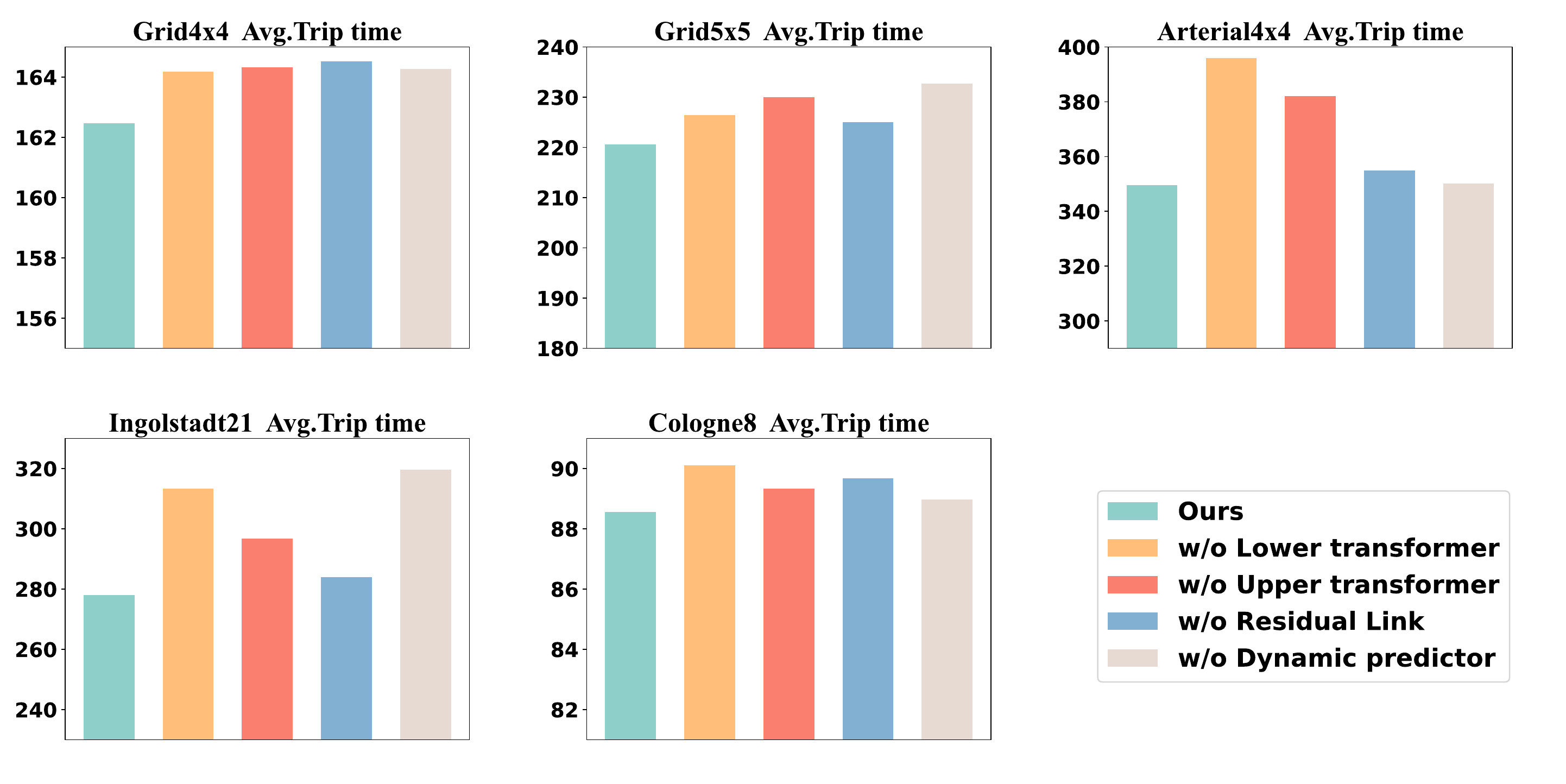} 
\caption{Ablation of each component in X-Light. The final solution attained the shortest trip time among all ablation experiments, showcasing the effectiveness of our design components.}
\label{fig:ablation}
\end{figure}

\begin{figure}[t]
\centering
\includegraphics[width=0.95\columnwidth]{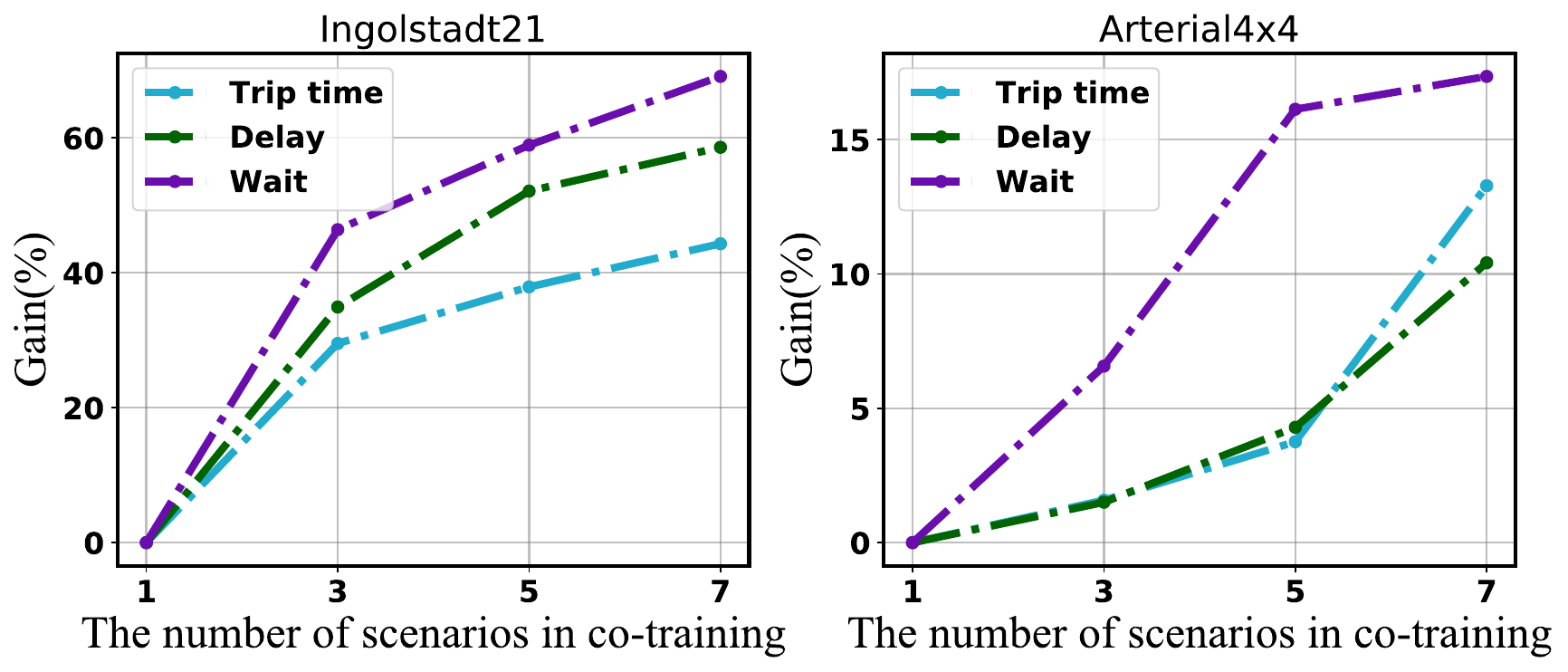} 
\caption{Impact of the number of scenarios co-training. As the number of scenarios increases, our performance also increases.}
\label{fig:ablation-num}
\end{figure}

\subsubsection{Enhanced Performance in Non-transfer Setting}
In this non-transfer setting, as shown in Table \ref{table:main-table},  the scenario in each column is present in training: specifically, all meta-learning-based methods from MetaLight to Ours are trained with all seven scenarios, while other methods from FTC to CoLight are trained and tested on the same one scenario. Figure~\ref{fig:duration_compare} further plots the top three methods' training process. 
\textbf{(1) TonT is also advantageous for non-transfer:} our method achieves the best results in four scenarios, with a \textbf{+6.20\%} gain on average and \textbf{+9.96\%} in \textit{Grid5$\times$5}. The gain is relatively lower than the Transfer setting, which is reasonable given we are dedicated to transfer. Besides, we have tiny \textbf{-1.02\%} loss compared with the best GESA in the \textit{Grid4$\times$4}. A potential reason is the over-simplicity of the scenario. \textbf{(2) Our multi-scenario co-training is more stable than MetaGAT:} As shown in Figure~\ref{fig:duration_compare}, our training is much more stable than MetaGAT (both are multi-agent setting), due to the co-training design and strong scenario-agnostic trained Upper Transformer. GESA is the most stable given it is single-agent and also in a multi-scenario co-training scheme like ours.


\subsubsection{Better Convergence of X-Light}
\begin{figure}[t]
\centering
\includegraphics[width=\columnwidth]{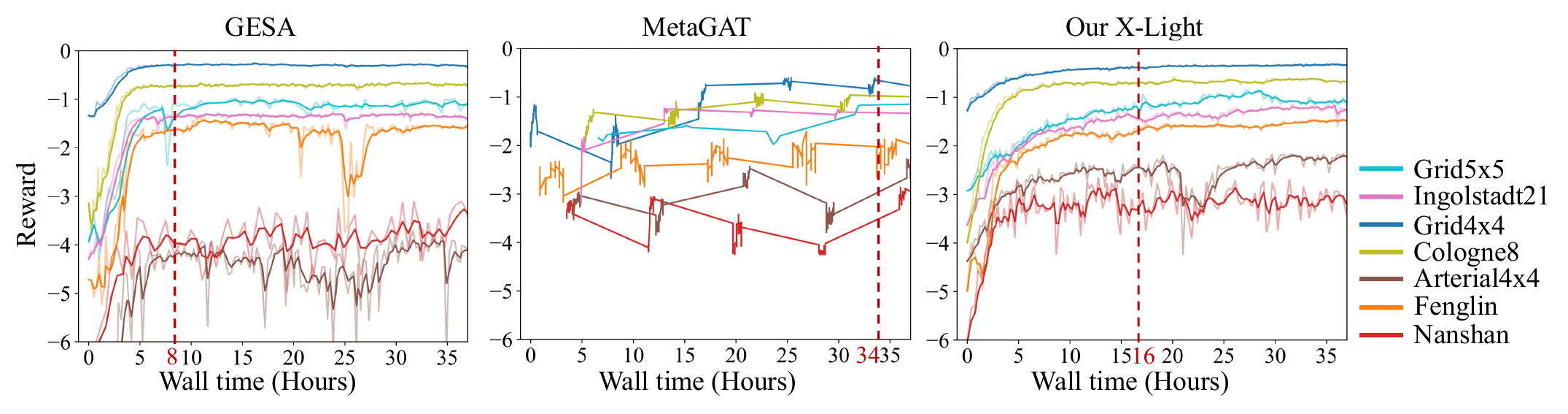} 
\caption{Training curve with wall times of the Top 3 methods. They utilize multi-scenarios during training. The red dotted line  approximates convergence time.}
\label{fig:walltime_compare}
\end{figure}

All the methods are trained for 37 hours and converged.  In Fig. \ref{fig:walltime_compare}, we show the Training curve with wall times of the Top 3 methods. While GESA only takes 8 hours to converge, it shows inferior performance after convergence because it is single-agent RL. In multi-agent RL, our X-Light is around twice as fast as MetaGAT (16 hours over 34 hours); this is because GNN is computationally heavy. 
Furthermore, our Multi-scenario Co-Training, where intersections are stochastically chosen from various scenarios within each batch, has also contributed to more stable and beautiful training curves than MetaGAT. 

\subsubsection{Ablation Studies}
We conduct ablations to verify the necessity of the Lower and Upper Transformers, residual link, and dynamic predictor.
\begin{itemize}
\setlength\itemsep{0.1em}
    \item \textbf{w/o Lower Transformer} replaces the Lower Transformer with MLP. No collaboration is formulated.
    \item\textbf{w/o Upper Transformer} replaces the Upper Transformer with GRU.
    \item \textbf{w/o Residual Link} removes the $\mathbf{o}_i^t$ in the actor input.
    \item \textbf{w/o Dynamic predictor} removes the dynamic predictor in the Upper Transformer, thus without $\mathcal{L}_{pred}$.
\end{itemize}
Figure~\ref{fig:ablation} shows the trip time results, with other metrics in Appx.~\ref{sub:appendix ablation studies}. 
We can conclude that (1) the Lower Transformer, which promotes collaboration, is the most critical component. (2) In the Upper Transformer, the Transformer model is better than GRU; moreover, the dynamic predictor is also very necessary to ensure the Upper Transformer learns scenario-diagnostic dynamics. (3) As discussed before, the residual link can further boost performance, especially in easier scenarios such as \textit{Grid4$\times$4} and \textit{Cologne8}, where the direct self-observation is more useful for decision in simple scenarios.

\textbf{Numbers of co-trained scenarios}: in Figure~\ref{fig:ablation-num}, we also conducted the impact of different numbers of co-training scenarios. We increase the number of co-trained scenarios from 1 to 7 and test in the two fixed scenarios (\textit{Ingolstadt21} and \textit{Arterial4$\times$4}). Our method consistently improves as the number of co-training scenarios increases. 

We also conducted experiments on the Upper Transformer, exploring various historical lengths, with details in Appx.~\ref{sub:appendix ablation studies}.

\begin{figure}[]
\centering
\includegraphics[width=0.95\columnwidth]{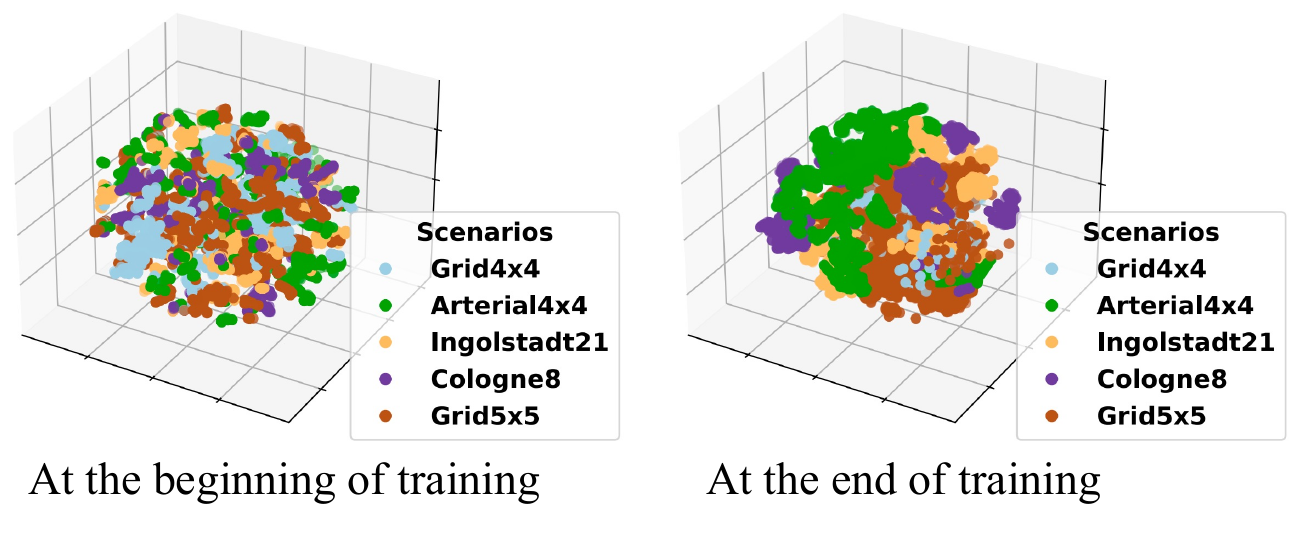} 
\caption{Visualizing Upper Transformer's output $\mathbf{z}^{\text{output}}$. Our model can group the decision dynamics from the same scenario together.}
\label{fig:Visualization}
\vspace{-10pt}
\end{figure}

\begin{figure}[t]
\centering
\includegraphics[width=1.04\columnwidth]{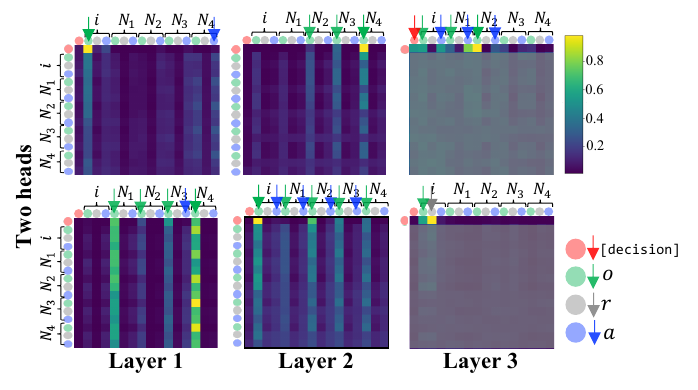} 
\caption{Visualizing Lower Transformer's attention (\textit{Cologne8}), with top attentions marked with $\downarrow$.  
Each row in attention is the weight of the corresponding token w.r.t all other tokens. We pick 2 heads over 3 layers. The last layer exclusively uses the \texttt{[decision]} token (only the first row is shown). Lower Transformer efficiently models these MDP features' interrelations: more attention to $o$ in shallow layers, and more to $a,r$ in deeper layers.
}
\label{fig:attention weights}
\vspace{-10pt}
\end{figure}

\subsubsection{Visualizing X-Light's Upper and Lower Transformers}
This visualization offers great insights on our TonT.

\textbf{Upper Transformer understands the scenario dynamics well}: We first analyze the output embeddings from the Upper Transformer. In Figure~\ref{fig:Visualization}, we visualize the model before and after training in five evaluation scenarios with the 3D t-SNE technique, and each scenario has a unique color. The Upper Transformer can map dynamics from the same scenario together, demonstrating the Upper Transformer's good understanding of various scenario dynamics.

\textbf{Lower Transformer collaborates well by fully utilizing $o,a,r$}: Furthermore, we visualize the attention weights of the Lower Transformers. In Figure~\ref{fig:attention weights}, we visualize attention weights for two heads across three layers in the \textit{Cologne8}. 
We find that (1) in the shallow Layer-1, attention is mainly to observations (green) of the target and its neighbors via different heads; in deeper Layer-2, more attention is to actions (blue); in deepest Layer-3, right the output being used for decision, all $o,r,a$, especially rewards (grey), have high attention (more cases in Appx.~\ref{sub:Preliminary:Visualization}); (2) This further justify the necessity of our Lower Transformer, with superiority of modeling $o,r,a$-interrelation over GNN-based collaborations. 

\section{Discussions}
In this section, we will discuss a few key questions the audience may be interested in.

\textbf{Discussion on Scalability:} 
Our model is scalable since it's decentralized: when it takes action, it only requires self-observation and observations of the nearest neighbor intersections. In reality, when controlling the traffic signals of a whole city, commonly the city is partitioned into multiple subareas according to the intersections' similarities in structure and traffic conditions \cite{guo2014dynamic}, with a subarea usually 4-7 intersections spatial spanning within 1 kilometer. This subarea control strategy is more effective and flexible when the traffic conditions differ between different districts.

\textbf{Discussion on compatibility of X-Light towards constraints on stages and cycle lengths:} Can our X-Light consider existing constraints on stages and cycle lengths, and deal with min-max green durations and pedestrian crossings? The answer is positive: our transformer-on-transformer (TonT) from X-Light is a general framework for transferring across cities; as long as the backbone RL-based TSC model considers the constraints, X-Light can also do it. In this work, we use FRAP as the RL backbone, which is an acyclic action that does not have a ``cycle'' of phases. When applying our method for cyclic control, we can change our backbone RL model to \cite{liang2019deep} (e.g., action is to +/- 5 seconds on a phase, or remain the same) and add maximum/minimum green durations or cycle length constraints (e.g., disable the action of +5s if when exceeding max-duration). 

\textbf{Discussion on transfer assumption for different regions} We will discuss from the two perspectives of junction and phase. \textbf{Junction-wise}: No specific requirements on the number of legs/movements: Since we use the GPI module, it maps various junctions into a unified structure, e.g., non-existing movements' states and actions will be zero-padded. n \textbf{Phase-wise}: Since we are acyclic control, we only need the common setting: each selected phase runs for 10s, and the yellow light is 3s.

\textbf{Discussion on the potential guarantees of the generated strategies} Firstly, the evaluation methodology of our zero-shot experiments validates the policies' generalization capabilities. 
Secondly, theoretical guarantees on performance lower bounds can be assured. The performance in unseen scenarios can be proven to exceed a polynomial function of the tasks' maximum mean discrepancy. The proof follows \cite{rostami2020using}.

\section{Conclusion} \label{sub Conclusions}

In this paper, we propose X-Light, an innovative framework that has two types of transformers with multi-scenario co-training. This design enhances the agent's transferability and enables collaboration across multiple intersections simultaneously. We use a large number of scenarios and comparison methods to verify the effectiveness of our proposed method. The outcomes highlight not only the exceptional performance of our method but also its robust transferability. 

For future work, we aim to transition our methodology into real-world applications, addressing the challenging sim2real problem. We hold the belief that this endeavor will significantly enhance urban traffic efficiency.



\bibliographystyle{named}
\bibliography{ijcai24}

\clearpage

\onecolumn
\renewcommand*\appendixpagename{\centering Appendices}
\begin{appendices}
\setcounter{figure}{0}
\setcounter{section}{0}
\renewcommand{\thefigure}{A\arabic{figure}}
\setcounter{table}{0}
\renewcommand{\thetable}{A\arabic{table}}
In Appx.~\ref{sub:Preliminary}, we will introduce the preliminary of this work. In Appx.~\ref{sub:more related}, we will present additional related work on multi-agent RL for TSC and applying two transformers in other fields. In Appx.~\ref{sub Implementation details}, we will provide details regarding the implementation details of X-Light. In Appx.~\ref{sub Detail statistics}, we will present more detailed statistics about the scenarios. In Appx.~\ref{sub:appendix wait time}, we introduce a new metric to evaluate each method and also demonstrate that our proposed method still state-of-the-art. In Appx.~\ref{sub:appendix ablation studies}, we document additional results from the ablation studies. Lastly, in Appx.~\ref{sub:Preliminary:Visualization}, we present more visualization attention weights of the Lower Transformer.

\section{Preliminary}\label{sub:Preliminary}
\subsection{Definitions}
\begin{figure}[ht]
\centering
\begin{minipage}{.5\textwidth}
\centering
\includegraphics[width=1.1\columnwidth]{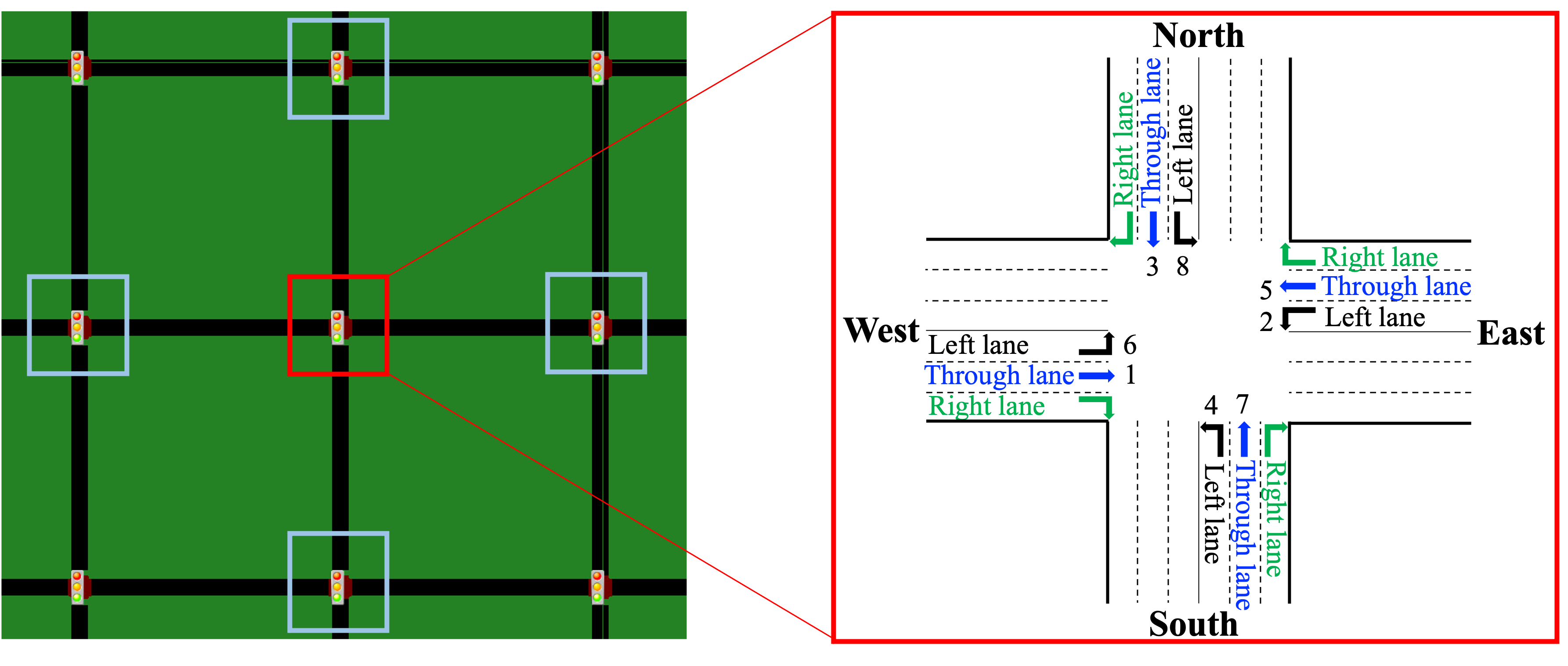} 
\caption{A simple scenario and a standard intersection structure. The red box is the target intersection and the blue box is its nearest top 4 neighbor intersections.}
\label{fig: Preliminary1}
\end{minipage}%
\begin{minipage}{0.5\textwidth}
\centering
\includegraphics[width=0.32\columnwidth]{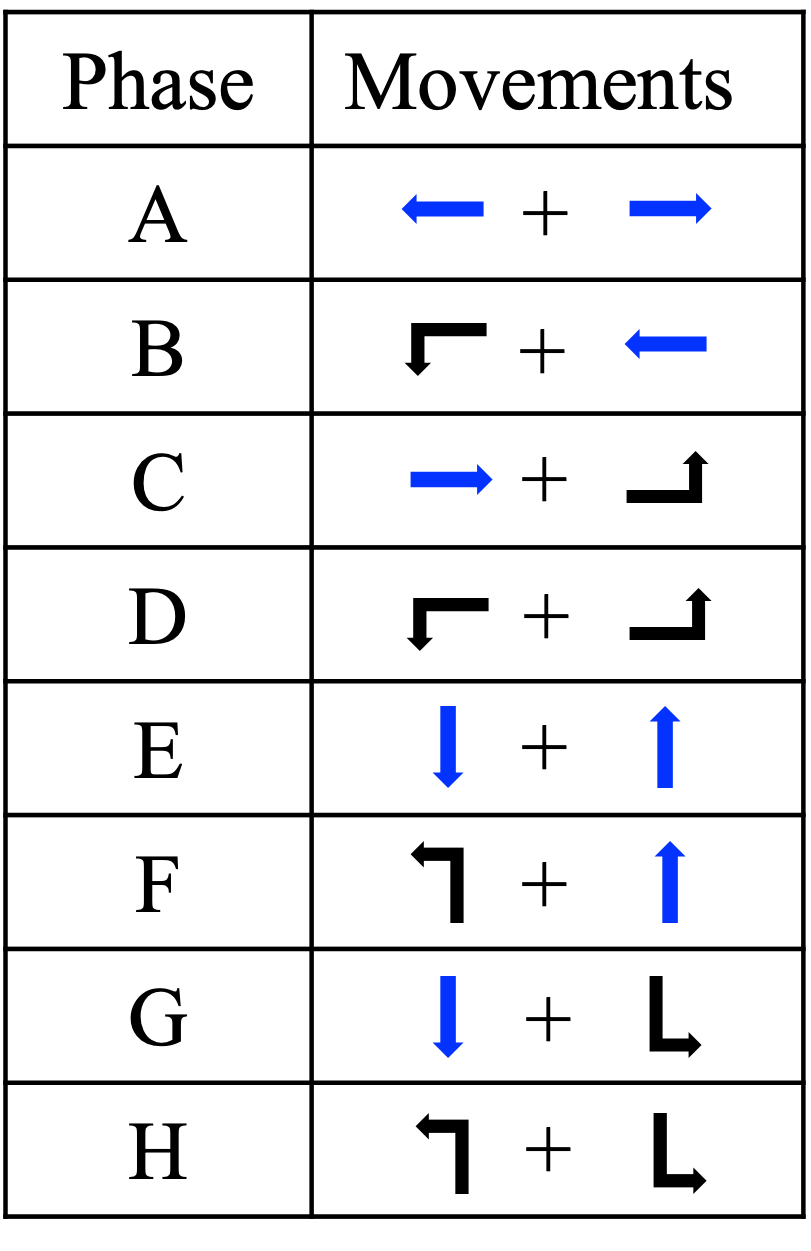} 
\caption{Enumerates eight typical signal phases}
\label{fig: Preliminary2}
\end{minipage}
\end{figure}

In this paper, we investigate traffic signal control of multi-intersection with different scenarios. To explain the basic concepts, we use a sample scenario with 9 intersections, and each intersection is a 4-arm intersection as shown in Figure~\ref{fig: Preliminary1}. The red box is the target intersection, the blue boxes indicate its neighbor intersections.

\textbf{Definition 1. \textit{Entering Lanes}:} For each intersection, the entering lanes indicate that vehicles can enter the intersection from these lanes. On the right of Figure~\ref{fig: Preliminary1}, a standard intersection has four approaches: north, east, south, and west, and each approach has three types of entering lanes: left lane, through lane, and right lane, therefore there are total 12 different movements for this intersection. In the real world, most intersections have four approaches, but some have 3-arm or 5-arm. 

\textbf{Definition 2. \textit{Phase}:} By selecting the execution phase of the next time step is a common way of traffic signal control. In the real world, we commonly do not control the left entering lanes~\cite{wolshon2016traffic}. As shown in Figure~\ref{fig: Preliminary2}, there are 8 typical signal phases in total, each of them consisting of two different and non-conflict movements. When a phase is executed, only the vehicles on the entering lanes of the phase can continue to drive.


\subsection{Problem Statement} \label{sub Problem Statement}

We define the problem of multi-agent traffic signal control in different scenarios as Meta-RL partially observed Markov decision processes (Meta-RL POMDPs)~\cite{humplik2019meta}. Given a set of scenarios $\mathcal{X}$ sampled over task distribution $\varepsilon$. Each agent controls one of the intersections $I_i$ from a scenario $x$, and each intersection has its top $n$ nearest neighboring intersections $\textit{\textbf{I}}_{\mathcal{N}_i}$, where $\mathcal{N}_i=\{N_1,\ldots,N_n\}$. Each agent observes part of the total system condition which only includes the transition information of the target intersection $i$ and its top $n$ nearest neighbor intersections. We can represent this as $\langle \mathcal{S},\mathcal{O},\mathcal{A},\mathcal{P}, r,\gamma \rangle$. $\mathcal{S}$ is the state space, and $s_t \in \mathcal{S}$ denotes the state at time step t. The observation is acquired through an observation function $D(s): \mathcal{S} \rightarrow \mathcal{O}$. $\mathcal{A}$ is the action space for each agent. At time step $t$, all agents in a scenario takes a joint action $a_t^{all} := \{\textbf{a}_t^1,...,\textbf{a}_t^{n_x}\}$, $n_x$ is the total number of the scenario $x$ and cause the state transition $\mathcal{P}(s_{t+1}|s_t, \textbf{a}_t^{all}) := \mathcal{S} \times \mathcal{A}^{n_x} \rightarrow \mathcal{S}$ and agent $i$ received a reward $r_t^i$.  Since the environment is partially observed, each agent only has access to its local observations, rewards and actions: $o_t^{\{i,\mathcal{N}_i\}}$, $a_{t-1}^{\{i,\mathcal{N}_i\}}$, and $r_{t-1}^{\{i,\mathcal{N}_i\}}$. When optimizing, we want to maximize cumulative reward for each agent: $\sum_t\gamma^tr_t^i$, where $\gamma$ is the discounted factor.

In meta-reinforcement learning, we need to be able to achieve great performance in a variety of scenarios, even if these scenarios have never been encountered before. As shown in Figure~\ref{fig: intro}, different surfaces represent different tasks, one goal is to learn $\theta$ to make a distinction among the tasks in the embedding space $\theta$, and minimize loss across different tasks:
\begin{equation}
   \theta = \arg\min_{\theta} \frac{1}{m}\sum_{i=1}^m\mathcal{L}(f_{\theta}(I_i), I_i)
\end{equation}
where $\theta$ is the policy parameter; $f_{\theta}(I_i)$ is the policy output, which denotes the policy to solve the task $I_i$; $\mathcal{L}$ is the loss.

\section{More Related Work}\label{sub:more related}
\subsection{Multi-Agent RL for TSC}
Multi-Agent RL-based TSC methods aim to optimize the efficiency of vehicle traffic within a given area by learning the collaboration among intersections. Early research in this area~\cite{prashanth2011reinforcement,van2016coordinated,xu2013study,prashanth2010reinforcement} through centralized optimization over joint actions among all intersections of a scenario. But this will cause two problems: 1. As the number of intersections increases, there will be a problem of dimensionality curse. 2. This approach does not generalize in meeting new scenarios. To counter these issues, independent or decentralized methods are proposed. One way~\cite{chen2020toward} is to guide cooperation by introducing the Pressure in the state and reward, which can reflect the intersection state associated with the upstream and downstream. Another way~\cite{el2013multiagent,nishi2018traffic,chu2019multi,wei2019colight,guo2021urban} is to include neighbor information into states or hidden states. Although they can all cooperate well with their neighbors, these methods are used in simple scenarios and the relationship between neighbors is static. For a new scenario, the state transitions between themselves and their neighbors are often quite different. 

\subsection{Applying Two Transformers in Other Fields.} 
In computer vision, some methods apply two Transformers to handle tasks like image classification and person search, e.g., TNT \cite{han2021transformer}., ViViT \cite{Arnab2021ViViT}, DualFormer \cite{Liang2022DualFormer} and COAT \cite{Yu2022CascadeTransformers}. However, our TonT model and these works are totally different.  For example, TNT’s outer Transformer takes the coarse-grain patches, and the inner one further takes the finer patches from the coarse patch. For our TonT, the lower Transformer is to perceive the environment dynamics and decides on collaboration: it reads the MDP including $o,a,r$ and uses attention to aggregate that information from the target and neighbor, and the upper transformer is to make the scenario-agnostic decision: it learns through the various scenarios decision dynamic and gain generalizability via memory reinstatement mechanism. As a result, TNT can only be used for vision tasks, but our TonT is designed for RL-based TSC tasks.

\section{Implementation details of X-Light}
\label{sub Implementation details}
In this section, we will give implementation details of X-Light. 

First, we will introduce how to get the state $o_t^i$. Initially, we utilize the GPI module proposed in GESA to reduce manual labeling and structured intersections with different lane counts and orientations. Subsequently, to obtain the state $o_t^i$, we concatenate each feature obtained from the MLPs. It can be formulated as below:
\begin{equation} \label{ot equation}
\begin{aligned}
    &\widetilde{o}_{t,b}^i=GPI(o_{t,b}^i), \\
    &o_t^i=\|^{B}_{b=1}Sigmoid(\text{MLP}_b(\widetilde{o}_{t,b}^i))
\end{aligned}
\end{equation}
where $B$ is the number of total state features, $Sigmoid$ is the activation function, and $\|$ denotes concatenation.

Furthermore, the training parameters and the weight of each reward component can be found in Table~\ref{table:Implementation details of TonTLight}.

\begin{table}[h!]
\small
\centering
\resizebox{0.35\columnwidth}{!}{%
\begin{tabular}{ll}
\hline
Items                                   & Details \\ \hline
Learning rate                           & 5e-4    \\
Actor loss coefficient $\alpha$                  & 1       \\
Critic loss coefficient $\beta$                 & 1       \\
Dynamic predictor loss coefficient $\kappa$      & 1e-2    \\
Entropy coefficient                     & 1e-2    \\
GAE $\lambda$                              & 0.95    \\
Discount factor $\gamma$                                   & 0.99    \\
Clipping $\epsilon$                         &0.2    \\ \hline
Number of Lower transformer layer      & 3       \\
Number of Lower transformer head        & 6       \\
Number of Upper transformer layer    & 3       \\
Number of Upper transformer head        & 6       \\
Lower transformer embedding dimension $d$ & 114     \\
Upper transformer embedding dimension $d'$ & 114     \\
Upper transformer horizon $K$               & 20      \\
The number of neighbor intersections $n$    & 4       \\ \hline
Weight of Queue length in Reward                            & -1e-3   \\
Weight of Wait time in Reward                              & -1e-3   \\
Weight of Delay time in Reward                             & -1e-5   \\
Weight of Pressure in Reward                               & -5e-3   \\ \hline
\end{tabular}
}
\caption{Implementation details of X-Light}
\label{table:Implementation details of TonTLight}
\end{table}

\section{The Detailed statistics about the scenarios}
\label{sub Detail statistics}

The important attributes of the scenarios are shown in Table \ref{tab: Data statistics of datasets}, including the total amount of intersections(Total Int.), the number of traffic flow types in each scenario(Flow type), and the amount of 2-arm to 4-arm intersections.

\begin{table}[h!]
\centering
\small
\resizebox{0.55\columnwidth}{!}{%
\begin{tabular}{llllllll}
\hline
Scenarios      & Country & Type & \#Total Int. &\#Flow types & \#2-arm & \#3-arm & \#4-arm \\ \hline
\textit{Grid $4 \times 4$}   & synthetic & region & 16   & 1400                      & 0          & 0          & 16         \\
\textit{Avenue $4 \times 4$}   & synthetic & region & 16  & 1400                    & 0          & 0          & 16         \\
\textit{Grid $5 \times 5$}     & synthetic & region & 25   & 2                      & 0          & 0          & 25         \\
\textit{Cologne8}     & Germany & region & 8           & 1               & 1          & 3          & 4          \\
\textit{Ingolstadt21} & Germany & region & 12        & 1                 & 0          & 8         & 4          \\
\textit{Fenglin}     & China & corridor & 7           & 1               & 0          & 2          & 5          \\
\textit{Nanshan} & China & region & 28        & 1                 & 1          & 6         & 21          \\ \hline
\end{tabular}
}
\caption{Statistics of the scenarios}
\label{tab: Data statistics of datasets}
\vspace{-5pt}
\end{table}

In actual traffic networks, numerous intersections deviate from the standard four-arm configuration, exhibiting varying numbers of lanes and orientations. To ensure the broad applicability of our approach across diverse scenarios, we deliberately conduct experiments within the contexts of two non-standard intersections: \textit{Cologne8} and \textit{Ingolstadt21}. 

These scenarios include not only typical four-arm intersections but also irregular intersections where the number and orientations of lanes deviate from the standard. Through experiments conducted in such settings, we demonstrate that our method not only effectively handles standard intersections, but also excels in managing these irregularities. This further emphasizes the robustness and potential wide applicability of our method.

\section{The main results on the evaluation metric of average waiting time}
\label{sub:appendix wait time}
Beside, two metric used in the main paper. We further add \textbf{average wait time} to compare various methods. The average waiting time represents how long each car was stopped. 

The performance of our method in terms of average waiting time for new scenarios(zero-shot transfer) is presented in the Table~\ref{table:wait time unseen}. It can be observed that in the metric of waiting time, our method also achieves the best results. The performance of various methods on average waiting time for non-transfer is shown in Table~\ref{table:wait time main}. The results are consistent with the conclusions of the main results on paper. Finally, considering all three metrics(Trip time, Delay time, and Wait time), our algorithm has improved by 8.11\% compared with the best result in the baselines.

\begin{table}[h!]
\small
\centering
\resizebox{0.85\textwidth}{!}{%
\begin{tabular}{c|lllll}
\hline
\multirow{2}{*}{\textbf{Methods}} & \multicolumn{5}{c}{\textbf{Avg. Wait Time (seconds)}}                                                                                                                                       \\ \cline{2-6} 
                         & \multicolumn{1}{c}{\textit{transfer to Grid4$\times$4}} & \multicolumn{1}{c}{\textit{transfer to Grid5$\times$5}} & \multicolumn{1}{c}{\textit{transfer to Arterial4$\times$4}} & \multicolumn{1}{c}{\textit{transfer to Ingolstadt21}} & \multicolumn{1}{c}{\textit{transfer to Cologne8}} \\ \hline
FTC                      & 66.12 \fontsize{8pt}{8pt}\selectfont{± 0.32}                       & 408.25 \fontsize{8pt}{8pt}\selectfont{± 7.09}                      & 599.10 \fontsize{8pt}{8pt}\selectfont{± 6.40}                          & 127.31 \fontsize{8pt}{8pt}\selectfont{± 20.06}                   & 35.30 \fontsize{8pt}{8pt}\selectfont{± 1.05}                  \\
MaxPressure              & 37.78 \fontsize{8pt}{8pt}\selectfont{± 0.65}                       & 116.85 \fontsize{8pt}{8pt}\selectfont{± 12.50}                      & 495.52 \fontsize{8pt}{8pt}\selectfont{± 10.52}                         & 184.97 \fontsize{8pt}{8pt}\selectfont{± 3.56}                    & 11.19 \fontsize{8pt}{8pt}\selectfont{± 0.44}                 \\ \hline
MetaLight                & 31.73 \fontsize{8pt}{8pt}\selectfont{± 0.49}                       & {\ul 106.36 \fontsize{8pt}{8pt}\selectfont{± 6.00}}                       & 212.63 \fontsize{8pt}{8pt}\selectfont{± 4.26}                          & 110.55 \fontsize{8pt}{8pt}\selectfont{± 5.03}                    & 8.52 \fontsize{8pt}{8pt}\selectfont{± 0.23}                  \\
GESA                     & 28.45 \fontsize{8pt}{8pt}\selectfont{± 0.72}                       & 135.22 \fontsize{8pt}{8pt}\selectfont{± 20.44}                     & 242.32 \fontsize{8pt}{8pt}\selectfont{± 2.72}                          & 135.07 \fontsize{8pt}{8pt}\selectfont{± 8.02}                    & {\ul 7.14 \fontsize{8pt}{8pt}\selectfont{± 0.19}}                  \\ \hline
MetaGAT                  & {\ul 26.46 \fontsize{8pt}{8pt}\selectfont{± 0.00}}                        & 120.72 \fontsize{8pt}{8pt}\selectfont{± 0.00}                       & {\ul 208.18 \fontsize{8pt}{8pt}\selectfont{± 0.00}}                          & {\ul 99.45 \fontsize{8pt}{8pt}\selectfont{± 0.00}}                     & 7.98 \fontsize{8pt}{8pt}\selectfont{± 0.00}                   \\ \cline{2-6} 
Ours                     & \textbf{24.71 \fontsize{8pt}{8pt}\selectfont{± 0.00}}              & \textbf{97.67 \fontsize{8pt}{8pt}\selectfont{± 3.31}}              & \textbf{206.45 \fontsize{8pt}{8pt}\selectfont{± 0.00}}                 & \textbf{94.17 \fontsize{8pt}{8pt}\selectfont{± 0.00}}            & \textbf{7.07 \fontsize{8pt}{8pt}\selectfont{± 0.00}}         \\ \hline
\end{tabular}}
\caption{Avg. wait time for transferring to unseen scenarios. The best \textbf{boldfaced} and second best \underline{underlined}. Our method also achieves the best.}
\label{table:wait time unseen}
\end{table}

\begin{table}[h!]
\small
\centering
\resizebox{0.75\textwidth}{!}{%
\begin{tabular}{c|lllll}
\hline
\multirow{2}{*}{\textbf{Methods}} & \multicolumn{5}{c}{\textbf{Avg. Wait Time (seconds)}}                                                                                                                  \\ \cline{2-6} 
                                  & \multicolumn{1}{c}{\textit{Grid4$\times$4 (seen)}}    & \multicolumn{1}{c}{\textit{Grid5$\times$5 (seen)}} & \multicolumn{1}{c}{\textit{Arterial4$\times$4 (seen)}}    & \multicolumn{1}{c}{\textit{Ingolstadt21 (seen)}}  & \multicolumn{1}{c}{\textit{Cologne8 (seen)}}   \\ \hline
FTC                               & 66.12 \fontsize{8pt}{8pt}\selectfont{± 0.32}                   & 408.25 \fontsize{8pt}{8pt}\selectfont{± 7.09}               & 599.10 \fontsize{8pt}{8pt}\selectfont{± 6.40}                        & 127.31 \fontsize{8pt}{8pt}\selectfont{± 20.06}                    & 35.30 \fontsize{8pt}{8pt}\selectfont{± 1.05}                    \\
MaxPressure                       & 37.78 \fontsize{8pt}{8pt}\selectfont{± 0.65}                   & 116.85 \fontsize{8pt}{8pt}\selectfont{± 12.50}               & 495.52 \fontsize{8pt}{8pt}\selectfont{± 10.52}                     & 184.97 \fontsize{8pt}{8pt}\selectfont{± 3.56}                     & 11.19 \fontsize{8pt}{8pt}\selectfont{± 0.44}                   \\ \hline
MPLight                           & 41.14 \fontsize{8pt}{8pt}\selectfont{± 0.79} & 106.82 \fontsize{8pt}{8pt}\selectfont{± 4.26}                & 349.69 \fontsize{8pt}{8pt}\selectfont{± 43.48} & 130.57 \fontsize{8pt}{8pt}\selectfont{± 9.31} & 12.06 \fontsize{8pt}{8pt}\selectfont{± 0.32} \\
IPPO                              & 29.59 \fontsize{8pt}{8pt}\selectfont{± 1.15}                   & 108.61 \fontsize{8pt}{8pt}\selectfont{± 8.75}               & 289.89 \fontsize{8pt}{8pt}\selectfont{± 25.53}                     & 185.88 \fontsize{8pt}{8pt}\selectfont{± 32.55}                    & 8.47 \fontsize{8pt}{8pt}\selectfont{± 0.20}                   \\
rMAPPO                            & 27.07 \fontsize{8pt}{8pt}\selectfont{± 0.54}                   & 153.38 \fontsize{8pt}{8pt}\selectfont{± 7.44}               & 407.25 \fontsize{8pt}{8pt}\selectfont{± 41.15}                     & 264.41 \fontsize{8pt}{8pt}\selectfont{± 28.36}                    & 14.01 \fontsize{8pt}{8pt}\selectfont{± 1.12}                   \\
CoLight                           & 25.56 \fontsize{8pt}{8pt}\selectfont{± 0.00}                   & 100.58 \fontsize{8pt}{8pt}\selectfont{± 0.00}               & 247.88 \fontsize{8pt}{8pt}\selectfont{± 0.00}                      & 147.91 \fontsize{8pt}{8pt}\selectfont{± 0.00}                     & 8.18 \fontsize{8pt}{8pt}\selectfont{± 0.00}                    \\ \hline
MetaLight                         & 31.79 \fontsize{8pt}{8pt}\selectfont{± 0.71}                   & {\ul 98.57 \fontsize{8pt}{8pt}\selectfont{± 5.32}}               & 241.88 \fontsize{8pt}{8pt}\selectfont{± 11.29}                     & 104.08 \fontsize{8pt}{8pt}\selectfont{± 3.65}                     & 8.44 \fontsize{8pt}{8pt}\selectfont{± 0.31}                    \\
GESA                              & \textbf{23.86 \fontsize{8pt}{8pt}\selectfont{± 0.58}}          & 102.98 \fontsize{8pt}{8pt}\selectfont{± 8.66}               & 234.78 \fontsize{8pt}{8pt}\selectfont{± 6.02}                      & 136.43 \fontsize{8pt}{8pt}\selectfont{± 5.47}                     & {\ul 7.59 \fontsize{8pt}{8pt}\selectfont{± 0.28}}                    \\ \hline
MetaGAT                           & 27.25 \fontsize{8pt}{8pt}\selectfont{± 0.00}                   & 123.78 \fontsize{8pt}{8pt}\selectfont{± 0.00}               & {\ul 219.36 \fontsize{8pt}{8pt}\selectfont{± 0.27}}                      & {\ul 103.35 \fontsize{8pt}{8pt}\selectfont{± 0.62}}                     & 8.18 \fontsize{8pt}{8pt}\selectfont{± 0.00}                    \\ \cline{2-6} 
Ours                              & {\ul 24.57 \fontsize{8pt}{8pt}\selectfont{± 0.00}}                   & \textbf{79.62 \fontsize{8pt}{8pt}\selectfont{± 0.00}}       & \textbf{214.21 \fontsize{8pt}{8pt}\selectfont{± 0.00}}             & \textbf{96.05 \fontsize{8pt}{8pt}\selectfont{± 0.00}}             & \textbf{6.75 \fontsize{8pt}{8pt}\selectfont{± 0.00}}           \\ \hline
\end{tabular}}
\caption{Avg. wait time for various scenarios observed during training. Our method is also advantageous for wait time.}
\label{table:wait time main}
\end{table}

\section{The detailed results of the ablation studies}
\label{sub:appendix ablation studies}
In this section, we will the detailed results of the ablation studies, and all of these contain all three metrics. 

\begin{table}[h!]
\centering
\small
\resizebox{0.75\textwidth}{!}{%
\begin{tabular}{c|c|lll|l}
\hline
\multicolumn{1}{l|}{\multirow{2}{*}{\# scenarios}} & \multicolumn{1}{c|}{\multirow{2}{*}{Chosen scenarios}} & \multicolumn{3}{c|}{Metrics}                                                                  & \multirow{2}{*}{\textbf{Improvement in \%}} \\ \cline{3-5}
\multicolumn{1}{l|}{}                              & \multicolumn{1}{l|}{}                                  & Trip Time                & Delay                    & Wait                     & \multicolumn{1}{l}{}                                \\ \hline
1                                                  & \textit{\textbf{Ingolstadt21}}                                  & 499.22  \fontsize{8pt}{8pt}\selectfont{± 8.02}                  &  387.37 \fontsize{8pt}{8pt}\selectfont{± 0.77}                  & 310.87 \fontsize{8pt}{8pt}\selectfont{± 5.43}                  &   -                              \\ \hline
\multirow{2}{*}{3}                                 & \textit{\textbf{Ingolstadt21}}                                  & \multirow{2}{*}{351.84 \fontsize{8pt}{8pt}\selectfont{± 0.00}}  & \multirow{2}{*}{252.09 \fontsize{8pt}{8pt}\selectfont{± 0.00}} & \multirow{2}{*}{166.67 \fontsize{8pt}{8pt}\selectfont{± 0.00}} &  \textbf{+36.94\%} w.r.t one scenario only                            \\
                                                   & +\textit{Cologne8, Arterial 4$\times$4}                                &                               &                               &                               &                                \\ \hline
\multirow{3}{*}{5}                                 & \textit{\textbf{Ingolstadt21}}                                  & \multirow{3}{*}{309.99 \fontsize{8pt}{8pt}\selectfont{± 0.00}} & \multirow{3}{*}{185.53 \fontsize{8pt}{8pt}\selectfont{± 0.00}} & \multirow{3}{*}{127.70 \fontsize{8pt}{8pt}\selectfont{± 0.00}} & \textbf{+49.64\%}  w.r.t one scenario only                                    \\
                                                   & +\textit{Cologne8, Arterial 4$\times$4}                                &                               &                               &                               &                                \\
                                                   & +\textit{Grid4$\times$4, Grid5$\times$5}                                      &                               &                               &                               &                                \\ \hline
\multirow{4}{*}{7}                                 & \textit{\textbf{Ingolstadt21}}                                  & \multirow{4}{*}{278.05 \fontsize{8pt}{8pt}\selectfont{± 0.00}} & \multirow{4}{*}{160.39 \fontsize{8pt}{8pt}\selectfont{± 0.00}} & \multirow{4}{*}{96.05 \fontsize{8pt}{8pt}\selectfont{± 0.00}} & \textbf{+57.33\%}  w.r.t one scenario only                                   \\
                                                   & +\textit{Cologne8, Arterial 4$\times$4}                                &                               &                               &                               &  \multicolumn{1}{l}{}                               \\
                                                   & +\textit{Grid4$\times$4, Grid5$\times$5}                                      &                               &                               &                               & \multicolumn{1}{l}{}                                \\
                                                   & +\textit{Fenglin, Nanshan}                                      &                               &                               &                               & \multicolumn{1}{l}{}                                \\ \hline
\end{tabular}%
}
\caption{Results of co-training with different numbers of scenarios based on \textit{Ingolstadt21}. }
\label{table:ablation-table3}
\end{table}

\begin{table}[h!]
\centering
\small
\resizebox{0.75\textwidth}{!}{%
\begin{tabular}{c|c|lll|l}
\hline
\multicolumn{1}{l|}{\multirow{2}{*}{\# scenarios}} & \multicolumn{1}{c|}{\multirow{2}{*}{Chosen scenarios}} & \multicolumn{3}{c|}{Metrics}                                                                  & \multirow{2}{*}{\textbf{Improvement in \%}} \\ \cline{3-5}
\multicolumn{1}{l|}{}                              & \multicolumn{1}{l|}{}                                  & Trip Time                & Delay                    & Wait                     & \multicolumn{1}{l}{}                                \\ \hline
1                                                  & \textit{\textbf{Arterial 4$\times$4}}                                  & 403.16 \fontsize{8pt}{8pt}\selectfont{± 0.00}                  & 778.93 \fontsize{8pt}{8pt}\selectfont{± 0.00}                  & 259.19 \fontsize{8pt}{8pt}\selectfont{± 0.00}                  &   -                              \\ \hline
\multirow{2}{*}{3}                                 & \textit{\textbf{Arterial 4$\times$4}}                                  & \multirow{2}{*}{396.80 \fontsize{8pt}{8pt}\selectfont{± 0.00}}  & \multirow{2}{*}{767.22 \fontsize{8pt}{8pt}\selectfont{± 0.00}} & \multirow{2}{*}{242.16 \fontsize{8pt}{8pt}\selectfont{± 0.00}} &  \textbf{+3.22\%} w.r.t one scenario only                             \\
                                                   & +\textit{Cologne8, Ingolstadt21}                                &                               &                               &                               &                                \\ \hline
\multirow{3}{*}{5}                                 & \textit{\textbf{Arterial 4$\times$4}}                                  & \multirow{3}{*}{388.02 \fontsize{8pt}{8pt}\selectfont{± 0.00}} & \multirow{3}{*}{745.49 \fontsize{8pt}{8pt}\selectfont{± 0.00}} & \multirow{3}{*}{217.38 \fontsize{8pt}{8pt}\selectfont{± 0.00}} &\textbf{+8.06\%}   w.r.t one scenario only                                   \\
                                                   & +\textit{Cologne8, Ingolstadt21}                                &                               &                               &                               &                                \\
                                                   & +\textit{Grid4$\times$4, Grid5$\times$5}                                      &                               &                               &                               &                                \\ \hline
\multirow{4}{*}{7}                                 & \textit{\textbf{Arterial 4$\times$4}}                                  & \multirow{4}{*}{349.60 \fontsize{8pt}{8pt}\selectfont{± 0.00}} & \multirow{4}{*}{697.79 \fontsize{8pt}{8pt}\selectfont{± 0.00}} & \multirow{4}{*}{214.21 \fontsize{8pt}{8pt}\selectfont{± 0.00}} & \textbf{+13.68\%}  w.r.t one scenario only                                  \\
                                                   & +\textit{Cologne8, Ingolstadt21}                                &                               &                               &                               &  \multicolumn{1}{l}{}                               \\
                                                   & +\textit{Grid4$\times$4, Grid5$\times$5}                                      &                               &                               &                               & \multicolumn{1}{l}{}                                \\
                                                   & +\textit{Fenglin, Nanshan}                                      &                               &                               &                               & \multicolumn{1}{l}{}                                \\ \hline
\end{tabular}%
}
\caption{Results of co-training with different numbers of scenarios based on \textit{Aterial 4$\times$4}. }
\label{table:ablation-table2}
\end{table}

Table~\ref{table:ablation-table3} and Table~\ref{table:ablation-table2} show the effect of increasing the number of co-training scenarios. we can see that as the number of co-training scenarios increases, the results of the three metrics and stability continue to improve. The increase in the number of scenarios enhances the diversity of data available to the model, thereby preventing the model from becoming trapped in a local optimum caused by data distribution bias~\cite{jiang2023a}.

\begin{table}[h!]
\small
\centering
\resizebox{0.9\textwidth}{!}{%
\begin{tabular}{ll|lllll}
\hline
\multicolumn{1}{l|}{Scenarios}                          & Metrics        & \multicolumn{1}{c}{Ours} & w/o Residual Link & w/o Dynamic Predictor &w/o Upper Transformer & w/o Lower Transformer\\ \hline
\multicolumn{1}{l|}{\multirow{3}{*}{\textit{Grid4$\times$4}}}      & Trip Time & 162.47 \fontsize{8pt}{8pt}\selectfont{± 0.00}             & 164.52 \fontsize{8pt}{8pt}\selectfont{± 0.00}          & 164.27 \fontsize{8pt}{8pt}\selectfont{± 0.00}   & 164.32 \fontsize{8pt}{8pt}\selectfont{± 0.00}       & 164.18 \fontsize{8pt}{8pt}\selectfont{± 0.00}          \\
\multicolumn{1}{l|}{}                              & Delay     & 50.27 \fontsize{8pt}{8pt}\selectfont{± 0.00}              & 52.79 \fontsize{8pt}{8pt}\selectfont{± 0.00}           & 52.44 \fontsize{8pt}{8pt}\selectfont{± 0.20}     & 51.58 \fontsize{8pt}{8pt}\selectfont{± 0.00}      & 52.42 \fontsize{8pt}{8pt}\selectfont{± 0.00}           \\
\multicolumn{1}{l|}{}                              & Wait      & 24.57 \fontsize{8pt}{8pt}\selectfont{± 0.00}              & 26.45 \fontsize{8pt}{8pt}\selectfont{± 0.00}           & 25.97 \fontsize{8pt}{8pt}\selectfont{± 0.00}    & 27.45 \fontsize{8pt}{8pt}\selectfont{± 0.00}       & 26.33 \fontsize{8pt}{8pt}\selectfont{± 0.00}           \\ \hline
\multicolumn{1}{l|}{\multirow{3}{*}{\textit{Grid5$\times$5}}}      & Trip Time & 220.63 \fontsize{8pt}{8pt}\selectfont{± 0.00}             & 225.05 \fontsize{8pt}{8pt}\selectfont{± 0.00}          & 232.75 \fontsize{8pt}{8pt}\selectfont{± 0.00}     & 229.96 \fontsize{8pt}{8pt}\selectfont{± 0.00}     & 226.43 \fontsize{8pt}{8pt}\selectfont{± 0.00}          \\
\multicolumn{1}{l|}{}                              & Delay     & 187.74 \fontsize{8pt}{8pt}\selectfont{± 0.00}             & 207.2 \fontsize{8pt}{8pt}\selectfont{± 0.00}           & 198.54 \fontsize{8pt}{8pt}\selectfont{± 0.00}    & 193.21 \fontsize{8pt}{8pt}\selectfont{± 0.00}      & 208.65 \fontsize{8pt}{8pt}\selectfont{± 0.00}          \\
\multicolumn{1}{l|}{}                              & Wait      & 79.62 \fontsize{8pt}{8pt}\selectfont{± 0.00}              & 82.32 \fontsize{8pt}{8pt}\selectfont{± 0.00}           & 90.46 \fontsize{8pt}{8pt}\selectfont{± 0.00}    & 79.95 \fontsize{8pt}{8pt}\selectfont{± 0.00}       & 84.89 \fontsize{8pt}{8pt}\selectfont{± 0.00}           \\ \hline
\multicolumn{1}{l|}{\multirow{3}{*}{\textit{Arterial4$\times$4}}}  & Trip Time & 349.60 \fontsize{8pt}{8pt}\selectfont{± 0.00}             & 354.99 \fontsize{8pt}{8pt}\selectfont{± 2.95}         & 350.11 \fontsize{8pt}{8pt}\selectfont{± 0.00}    & 382.14 \fontsize{8pt}{8pt}\selectfont{± 2.54}      & 395.90 \fontsize{8pt}{8pt}\selectfont{± 0.00}           \\
\multicolumn{1}{l|}{}                              & Delay     & 697.79 \fontsize{8pt}{8pt}\selectfont{± 0.00}             & 768.57 \fontsize{8pt}{8pt}\selectfont{± 14.47}        & 739.1 \fontsize{8pt}{8pt}\selectfont{± 0.00}       & 723.87 \fontsize{8pt}{8pt}\selectfont{± 0.00}     & 785.72 \fontsize{8pt}{8pt}\selectfont{± 0.00}        \\
\multicolumn{1}{l|}{}                              & Wait      & 214.21 \fontsize{8pt}{8pt}\selectfont{± 0.00}             & 201.22 \fontsize{8pt}{8pt}\selectfont{± 0.15}         & 185.4 \fontsize{8pt}{8pt}\selectfont{± 0.00}     & 206.38 \fontsize{8pt}{8pt}\selectfont{± 0.0}      & 223.02 \fontsize{8pt}{8pt}\selectfont{± 0.0}          \\ \hline
\multicolumn{1}{l|}{\multirow{3}{*}{\textit{Ingolstadt21}}} & Trip Time & 278.05 \fontsize{8pt}{8pt}\selectfont{± 0.00}             & 283.89 \fontsize{8pt}{8pt}\selectfont{± 0.00}          & 319.61 \fontsize{8pt}{8pt}\selectfont{± 17.09}    & 296.74 \fontsize{8pt}{8pt}\selectfont{± 0.00}    & 313.39 \fontsize{8pt}{8pt}\selectfont{± 0.00}          \\
\multicolumn{1}{l|}{}                              & Delay     & 160.39 \fontsize{8pt}{8pt}\selectfont{± 0.00}             & 176.22 \fontsize{8pt}{8pt}\selectfont{± 0.00}          & 196.56 \fontsize{8pt}{8pt}\selectfont{± 0.00}    & 172.63 \fontsize{8pt}{8pt}\selectfont{± 0.00}      & 202.80\fontsize{8pt}{8pt}\selectfont{± 0.00}           \\
\multicolumn{1}{l|}{}                              & Wait      & 96.05 \fontsize{8pt}{8pt}\selectfont{± 0.00}              & 99.45 \fontsize{8pt}{8pt}\selectfont{± 0.00}           & 128.6 \fontsize{8pt}{8pt}\selectfont{± 15.56}    & 105.25 \fontsize{8pt}{8pt}\selectfont{± 1.21}     & 125.82 \fontsize{8pt}{8pt}\selectfont{± 0.00}          \\ \hline
\multicolumn{1}{l|}{\multirow{3}{*}{\textit{Cologne8}}}     & Trip Time & 88.55 \fontsize{8pt}{8pt}\selectfont{± 0.00}              & 89.67 \fontsize{8pt}{8pt}\selectfont{± 0.00}           & 88.98 \fontsize{8pt}{8pt}\selectfont{± 0.00}      & 89.33 \fontsize{8pt}{8pt}\selectfont{± 0.00}     & 90.11 \fontsize{8pt}{8pt}\selectfont{± 0.00}           \\
\multicolumn{1}{l|}{}                              & Delay     & 24.31 \fontsize{8pt}{8pt}\selectfont{± 0.00}              & 25.40 \fontsize{8pt}{8pt}\selectfont{± 0.00}            & 24.91 \fontsize{8pt}{8pt}\selectfont{± 0.00}     & 25.62 \fontsize{8pt}{8pt}\selectfont{± 0.22}      & 26.36 \fontsize{8pt}{8pt}\selectfont{± 0.00}          \\
\multicolumn{1}{l|}{}                              & Wait      & 6.75 \fontsize{8pt}{8pt}\selectfont{± 0.00}               & 7.05 \fontsize{8pt}{8pt}\selectfont{± 0.28}           & 7.01 \fontsize{8pt}{8pt}\selectfont{± 0.00}      & 6.83 \fontsize{8pt}{8pt}\selectfont{± 0.14}      & 7.61 \fontsize{8pt}{8pt}\selectfont{± 0.00}           \\ \hline
\multicolumn{2}{l|}{Avg. Decline ratio}                             & \multicolumn{1}{c}{-}    & 4.07\%                & 7.12\%       & 4.25\%         & 10.39\%                \\ \hline
\end{tabular}
}
\caption{The detailed results of ablation studies reveal that each component contributes to improvement, showcasing the effectiveness of our design components.}
\label{table:ablation-table}
\end{table}

Table~\ref{table:ablation-table} shows the impact of important components in our proposed model, we found that each of these key components brought a performance boost to our method.

Table~\ref{table:ablation-table-length} illustrates the impact of various history lengths of the Upper Transformer on the results. A noteworthy enhancement in results occurs from K=10 to K=20. Furthermore, an improvement is observed in scenarios featuring complex and large traffic flow (\textit{Grid 5 $\times$ 5} and \textit{Arterial 4 $\times$ 4}) when transitioning from K=20 to K=30, whereas simpler scenarios show relatively minor changes. We also measured the average inference time for each model in \textit{Grid5 $\times$ 5}. We use GPU(NVIDIA A800) for inference, in each time step $\text{Ours}_{10}$ costs 0.0208s, $\text{Ours}_{20}$ costs 0.0230s, $\text{Ours}_{30}$ costs 0.0255s.

\begin{table}[h!]
\centering
\small
\resizebox{0.9\textwidth}{!}{%
\begin{tabular}{c|lllll|lllll}
\toprule
\multirow{2}{*}{Methods} & \multicolumn{5}{c|}{Avg. Trip Time (seconds)}                                                          & \multicolumn{5}{c}{Avg. Delay Time (seconds)}                                                                \\ \cline{2-11} 
                         & \multicolumn{1}{c}{Grid4$\times$4}                                                                       & \multicolumn{1}{c}{Grid5$\times$5}                                                                          & \multicolumn{1}{c}{Arterial4$\times$4}                                                                      & \multicolumn{1}{c}{Ingolstadt21}                                                                            & \multicolumn{1}{c|}{Cologne8}                                                                              & \multicolumn{1}{c}{Grid4$\times$4}                                                                      & \multicolumn{1}{c}{Grid5$\times$5}                                                                          & \multicolumn{1}{c}{Arterial4$\times$4}                                                                      & \multicolumn{1}{c}{Ingolstadt21}                                                                            & \multicolumn{1}{c}{Cologne8}                                                                               \\ \hline
Ours$_{10}$             & 162.71 \fontsize{8pt}{8pt}\selectfont{± 0.00}                        & 223.12 \fontsize{8pt}{8pt}\selectfont{± 0.00}                           & 374.85 \fontsize{8pt}{8pt}\selectfont{± 0.87}    &  283.02 \fontsize{8pt}{8pt}\selectfont{± 0.45}    & 88.61 \fontsize{8pt}{8pt}\selectfont{± 0.00}                           & {\ul 50.69 \fontsize{8pt}{8pt}\selectfont{± 0.00}}                        & 188.74 \fontsize{8pt}{8pt}\selectfont{± 0.00}                           & 725.06 \fontsize{8pt}{8pt}\selectfont{± 0.00}                           & 169.59 \fontsize{8pt}{8pt}\selectfont{± 2.37}                           & 24.41 \fontsize{8pt}{8pt}\selectfont{± 0.00}                           \\ \cline{2-11} 
Ours$_{20}$             & \textbf{162.47 \fontsize{8pt}{8pt}\selectfont{± 0.00}} & {\ul 220.63 \fontsize{8pt}{8pt}\selectfont{± 0.00}} & {\ul 349.60\fontsize{8pt}{8pt}\selectfont{ ± 0.00}} & {\ul 278.05 \fontsize{8pt}{8pt}\selectfont{± 0.00}} & \textbf{88.55 \fontsize{8pt}{8pt}\selectfont{± 0.00}} & \textbf{50.27 \fontsize{8pt}{8pt}\selectfont{± 0.00}} & {\ul 187.74 \fontsize{8pt}{8pt}\selectfont{± 0.00}} & {\ul 697.79\fontsize{8pt}{8pt}\selectfont{± 0.00}} & \textbf{160.39 \fontsize{8pt}{8pt}\selectfont{± 0.00}} & \textbf{24.31 \fontsize{8pt}{8pt}\selectfont{± 0.00}} \\ \cline{2-11}
Ours$_{30}$             & {\ul 162.57 \fontsize{8pt}{8pt}\selectfont{± 0.00}} & \textbf{219.7 \fontsize{8pt}{8pt}\selectfont{± 0.00}}  & \textbf{345.88\fontsize{8pt}{8pt}\selectfont{ ± 0.00}} & \textbf{276.06 \fontsize{8pt}{8pt}\selectfont{± 0.00}} & {\ul 88.58 \fontsize{8pt}{8pt}\selectfont{± 0.00}} & 50.92 \fontsize{8pt}{8pt}\selectfont{± 0.00} & \textbf{181.21 \fontsize{8pt}{8pt}\selectfont{± 0.00}} & \textbf{517.25\fontsize{8pt}{8pt} \selectfont{± 0.00}} & {\ul162.79 \fontsize{8pt}{8pt}\selectfont{± 0.00}} & {\ul 24.39 \fontsize{8pt}{8pt}\selectfont{± 0.00}} \\ \bottomrule
\end{tabular}
}
\caption{The ablation studies of different history lengths of Upper Transformer. A noteworthy enhancement in results occurs from K=10 to K=20. In complex scenarios, K=30 can continue to bring improvements, whereas simpler scenarios show relatively minor changes.}
\label{table:ablation-table-length}
\end{table}


\section{More visualization attention weights of the
Lower Transformer} \label{sub:Preliminary:Visualization}

\begin{figure}[h!]
\centering
\includegraphics[width=\columnwidth]{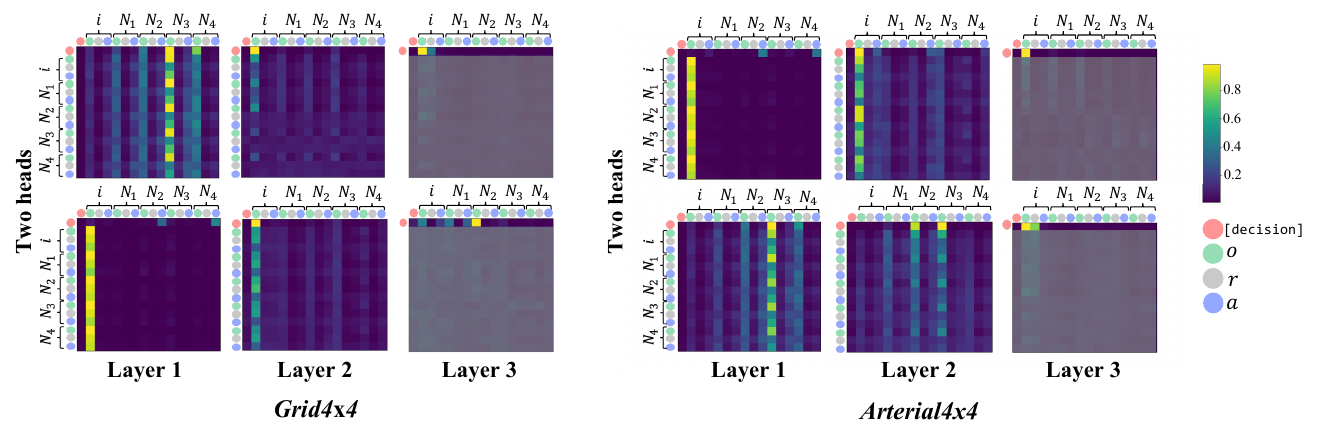} 
\caption{Visualizing Lower Transformer’s attention in \textit{Grid $4 \times 4$} and \textit{Arterial $4 \times 4$} scenarios.}
\label{fig:attention_app}
\end{figure}

\end{appendices}

\end{document}